\pdfoutput=1

\documentclass[a4paper,fleqn]{cas-sc}
\usepackage[numbers]{natbib}
\usepackage{lmodern}
\usepackage[T1]{fontenc}
\usepackage{amsmath,amssymb,amsfonts}
\usepackage{algorithmic}
\usepackage{algorithm}
\usepackage{array}
\usepackage[caption=false,font=normalsize,labelfont=sf,textfont=sf]{subfig}
\usepackage{textcomp}
\usepackage{stfloats}
\usepackage{url}
\usepackage{graphicx}
\usepackage{cite}
\usepackage[normalem]{ulem} 
\usepackage{threeparttable}
\usepackage{booktabs}
\usepackage{multirow}
\usepackage{graphicx}
\usepackage{tabularx} 
\usepackage{threeparttable}%

\usepackage{color}         
\usepackage{amsthm} 
\usepackage{booktabs} 
\usepackage{xcolor}
\usepackage{xspace}
\usepackage{pifont}
\newcommand{\stand}{STAND\xspace}
\newcommand{\myurl}{https://emorzz1g.github.io/STAND}

\usepackage[final,commandnameprefix=ifneeded]{changes}
\makeatletter
\renewcommand{\Changes@output}[7]{%
  \IfStrEq{#1}{deleted}{}{#3}%
}
\makeatother

\usepackage{xr}
\makeatletter
\let\orig@bibcite\bibcite
\def\bibcite#1#2{}
\let\bibcite\orig@bibcite
\makeatother

\newif\ifarxiv
\arxivfalse 

\makeatletter
\AtBeginDocument{%
  \ifdefined\@abspage@last
    \let\lastpage\@abspage@last
  \fi
}
\makeatother

\def\tsc#1{\csdef{#1}{\textsc{\lowercase{#1}}\xspace}}
\tsc{WGM}
\tsc{QE}
\tsc{EP}
\tsc{PMS}
\tsc{BEC}
\tsc{DE}

\begin{document}
\let\WriteBookmarks\relax
\def\floatpagepagefraction{1}
\def\textpagefraction{.001}
\shorttitle{Rethinking the Unsupervised Paradigm in TSAD}
\shortauthors{Z. Zhong et~al.}

\title [mode = title]{Rethinking the Unsupervised Paradigm in Time Series Anomaly Detection: An Empirical Study and Minimalist Baseline}                      



\author[1,2]{Zhijie Zhong}

\author[2,1]{Zhiwen Yu}
\cormark[1]
\ead{zhwyu@scut.edu.cn}

\author[2]{Kaixiang Yang}

\author[1]{Yongheng Liu}

\author[1]{Jun Jiang}

\author[2]{C. L. Philip Chen}

\affiliation[1]{organization={Pengcheng Laboratory},
                city={Shenzhen},
                postcode={518066},
                state={Guangdong},
                country={China}}
\affiliation[2]{organization={South China University of Technology},
                city={Guangzhou},
                postcode={510650},
                state={Guangdong},
                country={China}}


\cortext[cor1]{Corresponding author. Email: zhwyu@scut.edu.cn. Tel.: +86-20-62893506. Fax: +86-20-39380288.}

\begin{abstract}
Time series anomaly detection (TSAD) research predominantly focuses on unsupervised architectures to model normal data dynamics under label scarcity. However, practical deployments frequently archive historical operational incidents, questioning whether continuous unsupervised architectural scaling remains the most cost-effective pursuit. In this paper, we empirically quantify the trade-offs between supervisory signals and model complexity. We introduce \stand, a deliberately minimalist supervised baseline based on bidirectional recurrent encoding. Across five benchmark datasets evaluated under a strictly disjoint out-of-sample temporal protocol, minimalist supervised models achieve competitive or superior accuracy relative to complex unsupervised alternatives. Furthermore, controlled label-scaling and few-shot evaluations confirm that these supervisory advantages persist under constrained historical event budgets, while \stand exhibits higher prediction certainty and sharper event delineation than reconstruction-based baselines. These findings encourage the TSAD community to complement algorithmic complexity with data-centric label utilization when annotations are accessible. The source code, benchmark datasets, and an extended Supplementary Document are publicly available at~\url{\myurl}.

\end{abstract}



\begin{keywords}
time series anomaly detection \sep big data analytics \sep data-centric AI \sep supervised learning \sep limited supervision 
\end{keywords}

\maketitle

\section{Introduction}
In modern industrial data analytics, the exponential growth of sensor networks and streaming telemetry has elevated Time Series Anomaly Detection (TSAD) into a critical diagnostic tool \citep{is1}. However, real-world monitoring environments exhibit intricate spatio-temporal dynamics that complicate automated diagnosis \citep{is3}. Paradoxically, this deluge of streaming data is coupled with a severe scarcity of high-quality anomaly annotations, as manual labeling remains prohibitively expensive \citep{TSB-UAD}. To circumvent this bottleneck, the prevailing research paradigm has predominantly focused on Unsupervised Time-series Anomaly Detection (UTAD), designing increasingly heavy architectures to implicitly map normal data distributions \citep{catch}.

\begin{figure}[pos=htb]
    \centering
    \includegraphics[width=0.55\linewidth]{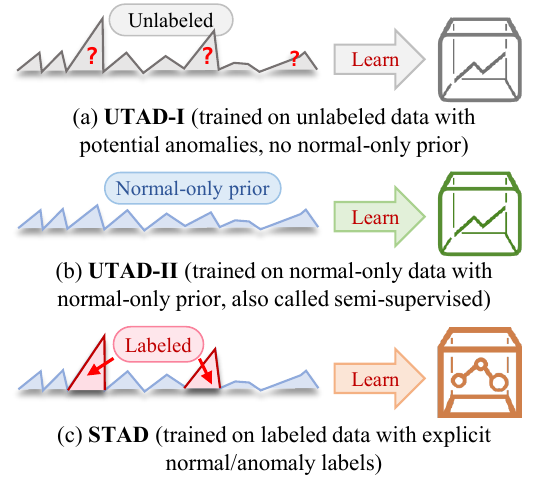}
    \caption{Classification of time series anomaly detection approaches.}
    \label{fig:fw}
\end{figure}

This unsupervised philosophy typically manifests in two standard methodologies. The first category, which we denote as UTAD Type I (UTAD-I), trains and infers directly on mixed, unlabeled streams that potentially contain contiguous anomalies (Figure \ref{fig:fw}(a)), a setting conventionally tackled by unsupervised projections such as PCA. The second, more dominant framework, termed UTAD Type II (UTAD-II), operates under a clean-data prior, optimizing models exclusively on anomaly-free historical segments to capture normal baseline behaviors (Figure \ref{fig:fw}(b)), as comprehensively surveyed in recent anomaly detection taxonomies \citep{ppt}. While the terminology varies, the foundational consensus governing both UTAD paradigms is strict: the objective function explicitly forbids the use of true anomaly labels during parameter optimization. 

Despite the academic dominance of the unsupervised paradigm, its core assumption is frequently challenged in actual industrial deployments, where a restricted budget of verified anomaly labels can be obtained through incident logs, operational alarms, or targeted expert auditing \citep{label1}, which has recently motivated emerging explorations into influence modeling and open-set supervision \citep{zhou2026impact}. Driven by an algorithm-centric focus, current literature largely overlooks the utility of these sparse supervisory signals, often pursuing performance gains primarily through architectural expansion. This trend motivates us to examine two practical questions: (i) \textit{In operational environments with limited annotations, how does the marginal gain from model complexity compare to that from sparse supervision? (ii) Can a minimalist supervised baseline provide an informative reference point for evaluating anomaly detection paradigms?}

\begin{figure}[pos=htb]
    \centering
    \includegraphics[width=\linewidth]{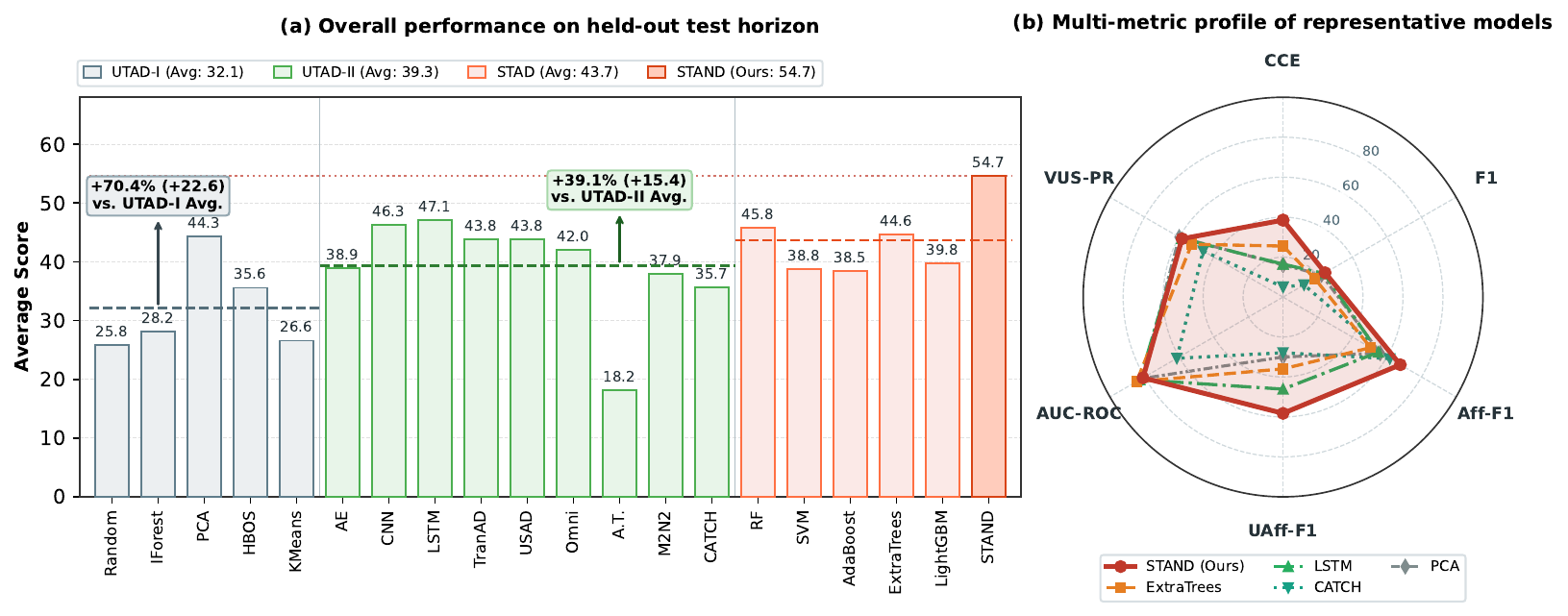}
    \caption{Performance comparison on the held-out test horizon across 20 benchmark methods and six evaluation metrics. (a) Mean performance across all datasets grouped by category, highlighting group averages and the substantial margins achieved by \stand over unsupervised baselines. (b) Multi-metric radar profile comparing representative top-tier methods from each category across individual evaluation dimensions.}
    \label{fig:comparison}
\end{figure}

To address these questions, rather than introducing another intricate architecture, we present a comprehensive benchmark alongside a deliberately minimalist \textbf{S}upervised \textbf{T}ime-series \textbf{AN}omaly \textbf{D}etection (\textbf{STAND}) baseline. By constructing \stand without mechanisms such as multi-head self-attention or generative diffusion modules, we isolate the impact of the supervisory signal. Our empirical evaluations and theoretical analyses demonstrate the substantial practical returns of supervision in established TSAD benchmarks. The core contributions of this work are summarized as follows:

\begin{itemize}
    \item \textbf{Quantifying the Empirical Returns of Supervision:} We demonstrate that under standard chronological out-of-sample evaluation, our minimalist supervised baseline (\stand) achieves competitive or superior detection accuracy compared to complex unsupervised architectures across five public datasets. As illustrated in Figure \ref{fig:comparison}(a), \stand attains an overall average score of $54.7$, exceeding the UTAD-I and UTAD-II category averages by $70.4\%$ and $39.1\%$, respectively. Furthermore, Figure \ref{fig:comparison}(b) shows that this advantage is prominent in prediction certainty (CCE) and unbiased event localization (UAff-F1), highlighting the high return on investment of targeted supervision on evaluated benchmarks.
    \item \textbf{Theoretical Intuition on Learning Difficulty:} Through the lens of statistical learning theory, we provide conceptual intuition contrasting the sample complexity of non-parametric density estimation against discriminative classification, explaining why direct supervision can yield substantial gains over purely unsupervised normal modeling in high-dimensional settings.
    \item \textbf{Prediction Certainty and Noise Robustness:} We show that while canonical UTAD methods frequently exhibit diffuse anomaly boundaries and higher uncertainty, explicit supervision anchors predictions effectively, maintaining structural and robust consistency even under asymmetric label noise.
    \item \textbf{A Unified Open-Source Benchmark:} Moving beyond fragmented evaluations, we introduce a unified evaluation framework and an open-source library\footnote{\url{\myurl}} that integrates supervised baselines into the TSAD evaluation ecosystem, facilitating balanced comparisons for future data-centric TSAD research.
\end{itemize}

\section{Related Work}\label{sec:related_work}
\subsection{Algorithm-Centric Time Series Anomaly Detection}\label{sec:algo_centric}

Early unsupervised Time Series Anomaly Detection (UTAD) methods were primarily based on classical machine learning techniques, such as Principal Component Analysis (PCA), Local Outlier Factor (LOF), and Isolation Forest (IForest). These methods transform time series data into static feature vectors using techniques like sliding window segmentation and statistical feature extraction, which are then fed into traditional anomaly detection algorithms. However, these approaches heavily depend on the quality of hand-crafted features, and their model structures are often insufficiently adapted to the dynamic nature of time series data.

The powerful capabilities of deep learning for automatic feature extraction and complex pattern modeling have been widely applied to UTAD. Existing algorithm-centric research in this area can be broadly classified into three main technical paths:

\textbf{Path I: Focusing on Model Architecture Improvement to Enhance Time Series Feature Extraction Capabilities.} This includes methods based on temporal convolutional and recurrent backbones \citep{tcn_ed}; Transformer-based approaches \citep{anom_trans}, which leverage attention mechanisms to improve long-range dependency modeling; state-space Mamba architectures \citep{s6_MemMambaAD}, which enhance the efficiency of processing long sequences; and Large Language Model (LLM)-guided frameworks \citep{llm_AnomalyLLM2024}, which utilize the knowledge learned by foundation models for anomaly recognition.

\textbf{Path II: Focusing on Establishing Anomaly Priors to Optimize Anomaly Determination.} Examples include deep one-class boundary estimation \citep{deep_svdd}, which maps normal instances to a minimal hypersphere; tree-based deep isolation ensembles \citep{deep_if}, which adapt partitioning concepts to high-dimensional representations; contrastive dissimilarity learning \citep{dcdetector}, which employs dual-attention representations to discern normal-abnormal discrepancies; and time-frequency decomposition models \citep{TFAD}, which transform time series into the frequency domain to capture multi-granularity anomalies.

\textbf{Path III: Exploring Limited and Imperfect Supervision to Alleviate Label Scarcity.} Beyond pure unsupervised formulations, a related trajectory explores limited supervision to alleviate annotation dependence. As cataloged in comprehensive benchmarks such as TSB-UAD \citep{TSB-UAD}, mainstream paradigms remain overwhelmingly unsupervised or one-class semi-supervised \citep{zhou2026impact}. Due to the practical difficulty of acquiring exhaustive labels, past studies have developed weakly supervised detection using multiple-instance learning with sequence-level tags, positive-unlabeled (PU) learning with surrogate risk estimators, active learning with human-in-the-loop query heuristics, and few-shot anomaly detection via meta-learning. In contrast to these approaches that introduce auxiliary surrogate objectives, transductive unlabeled access, or interactive query loops, \stand investigates standard sequence discrimination under an inductive out-of-sample regime with localized timestamp labels, establishing a straightforward baseline that isolates the empirical value of direct temporal supervision (with an extended literature survey and comparative discussion provided in Supplementary Section~\ref{app:limited_supervision}\footnote{Due to space constraints, all Supplementary Sections and Tables referenced in this paper are compiled into a standalone Supplementary Document, available at \url{\myurl}.}).

These approaches focus primarily on algorithmic and architectural design to mitigate label scarcity, leaving the practical utility of sparse supervisory signals largely unaddressed from a data perspective.

\subsection{Data-Centric Time Series Anomaly Detection}

Unlike algorithm-centric approaches, data-centric research revolves around optimizing the data layer to compensate for the deficiency of labels in unsupervised settings. However, due to the low prevalence of anomaly events and the high cost of labeling in most practical engineering scenarios, comprehensive surveys highlight that supervised time series anomaly detection remains sparsely explored in mainstream literature \citep{sv_landauerTimeSeriesAnalysis2018}, with only preliminary attempts utilizing multi-stage statistical classifiers under full labels \citep{STSAD}. Based on this constraint, past data-centric research on time series anomaly detection has mainly explored two optimization paths:

\textbf{Path I: Pseudo-Sample-Based Data Augmentation Strategies to Expand Data Value with Limited Information.} The core of this path is to substitute real labels with data transformations, creating pseudo-anomaly samples or task-oriented samples to provide learning signals for the model. This is implemented in two ways: First, \textit{unsupervised pseudo-sample generation}, where normal samples are modified into synthetic anomalies by simulating physical degradation, such as cut-and-paste transformations \citep{CutAddPaste}, self-supervised anomaly injection in Transformer encoders \citep{trs_jeongAnomalyBERTSelfSupervisedTransformer2023}, or calibrated margin generation \citep{couta}. Second, \textit{construction of task-guided samples}, where pseudo-boundary samples are synthesized via memory-augmented initialization \citep{adamembls}, or constructed through dissimilarity-based proxy bounds \citep{simad}. In addition, contextual learning architectures such as NCAD \citep{ncad} have explored conditioning predictions on reference windows under mixed or weak supervision, yet they still rely on intricate contextual encoders rather than examining the direct marginal efficiency of explicit anomaly labels.

\textbf{Path II: New Sample Generation Strategies Based on Distribution Modeling, Guiding Pattern Learning through Data Generation.} This path utilizes generative models (\textit{e.g.}, GANs, VAEs, Diffusion Models) to learn the distribution characteristics of normal time series data, thereby generating new samples that conform to realistic patterns and reinforcing the model's perception of normal modes, as exemplified by denoising diffusion models \citep{app_liDiffTADDenoisingDiffusion2024a}. The generated samples can directly augment the training set, mitigating data scarcity. Simultaneously, an anomaly determination baseline is established through the detection of deviations from the learned normal distribution, forming a closed loop between data generation and anomaly detection.

Although these studies explore performance improvements via data quality enhancement, their focus remains restricted to the unsupervised framework. They essentially bypass the reliance on true labels through pseudo-samples or generated samples, which may not fully exploit the few anomaly labels that are practically available. This represents a shared constraint of both the algorithm-centric and current data-centric approaches: both inherently assume a completely label-free prerequisite, which can leave the potential performance gain from limited labeling underexplored.

\begin{figure}[pos=htb]
    \centering
    \includegraphics[width=0.45\linewidth]{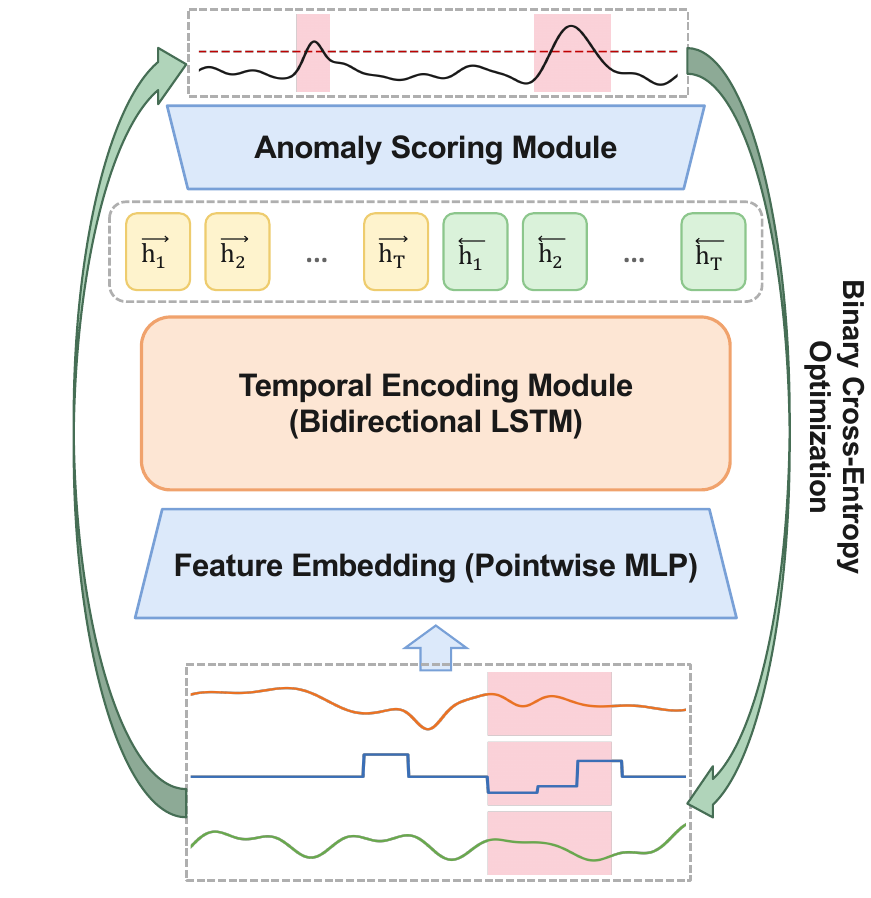}
    \caption{The overall architecture of the \stand~ framework. An input time series is first projected into latent space by the feature embedding. The resulting latent sequence is then fed into the temporal encoding module, which captures bidirectional temporal context. Finally, an anomaly scoring module outputs anomaly scores for each time step.}
    \label{fig:stand}
\end{figure}

\section{Methodology}
This section formalizes the time series anomaly detection problem under different label budgets and delineates the minimalist architecture of the proposed \stand~framework.

\subsection{Problem Formulation and Paradigms}
Let the multivariate time series be denoted as $\mathbf{X} \in \mathbb{R}^{T\times C}$, where $T$ is the temporal length and $C$ represents the number of feature channels. The corresponding step-wise ground-truth labels are defined as $\mathbf{y} \in \{0, 1\}^T$, where $y_t=1$ indicates an anomaly at time step $t$. The objective across all paradigms is to learn a mapping function that produces a continuous anomaly score sequence $\mathbf{S} \in \mathbb{R}^T$. 

Unsupervised vs. Supervised Paradigms: The main difference between UTAD and STAD lies not in the architecture, but in whether anomaly labels $\mathbf{y}$ are available during training. UTAD methods operate with zero anomaly labels ($\mathcal{D}_{\text{trn}} = \{ \mathbf{X}_i \}_{i=1}^{N_{\text{trn}}}$), requiring models to characterize normal data distributions and flag deviations. In contrast, STAD utilizes a limited labeling budget ($\mathcal{D}_{\text{trn}} = \{ (\mathbf{X}_i, \mathbf{y}_i) \}_{i=1}^{N_{\text{trn}}}$) to optimize parameter weights $\theta$ of a discriminative mapping $f_\theta(\mathbf{X})$ via empirical loss minimization. Rather than dismissing unsupervised approaches, this work investigates whether obtaining limited supervisory signals can provide a more cost-effective alternative to architectural expansion in practical deployments.

\subsection{The \stand Framework}\label{sec:proposed_method}
To separate the value of supervision from architectural complexity, \stand is built with simple, lightweight components (Figure \ref{fig:stand}). By avoiding complex self-attention or generative components, we ensure that performance differences can be attributed primarily to the supervisory signal rather than architectural over-engineering.

\subsubsection{Feature Embedding}
Raw inputs $x_t \in \mathbb{R}^C$ frequently reside in a space suboptimal for linear separation. To address this, we project the inputs into a discriminative $d_{\text{model}}$-dimensional latent space using a standard multi-layer perceptron (MLP). Specifically, we utilize linear transformations interleaved with Gaussian Error Linear Unit (GELU) activations \citep{gelu} and Layer Normalization \citep{layernorm} to stabilize the initial latent representations $\mathbf{H}^e \in \mathbb{R}^{T \times d_{\text{model}}}$.

\subsubsection{Temporal Encoding Module}
To capture temporal context without the quadratic cost of self-attention, the embedding sequence $\mathbf{H}^e$ is processed by a Bidirectional Long Short-Term Memory (BiLSTM) network. This bidirectional structure is effective for offline diagnostic auditing, as an anomalous event is often contextualized by both its onset and its subsequent recovery pattern. (For online streaming scenarios where future observations cannot be accessed, a causal forward-only variant is also evaluated in Section \ref{sec:component_ablation}.) The encoder produces hidden states $h_t^{\text{enc}} \in \mathbb{R}^{2d_{\text{model}}}$ by concatenating the forward and backward representations:
\begin{equation}h_t^{\text{enc}} = [\text{LSTM}_{\text{fwd}}(h_t^e); \text{LSTM}_{\text{bwd}}(h_t^e)]\label{eq:enc_bidir}\end{equation}

\subsubsection{Training Objective and Scoring}
The framework is trained end-to-end as a sequence-to-sequence binary classification task, avoiding the complex multi-stage training pipelines common in generative UTAD models. A linear projection maps each contextualized state $h_t^{\text{enc}}$ to a raw anomaly logit $s_t$. We optimize the network parameters via the Adam optimizer \citep{adam} by minimizing the standard timestamp-level Binary Cross-Entropy (BCE) loss:
\begin{equation}\mathcal{L}(\mathbf{S}, \mathbf{y}) = - \frac{1}{T} \sum_{t=1}^{T} \left[ y_t \log(\sigma(s_t)) + (1 - y_t) \log(1 - \sigma(s_t)) \right]\end{equation}
where $\sigma(\cdot)$ denotes the sigmoid activation function used to compute the prediction probabilities. We optimize this objective using the standard unweighted BCE loss (i.e., with positive class weight $w_{\text{pos}}=1$). As evaluated in Supplementary Section~\ref{app:imbalance}, introducing positive class re-weighting under severe imbalance (e.g., Water with $1.05\%$ anomalies) introduces hyperparameter sensitivity without providing consistent performance gains (Aff-F1 remaining within $90.78$--$92.21$), making the unweighted formulation the most parsimonious choice.

\section{Theoretical Analysis}
\label{sec:theoretical_analysis}

To better understand the properties of \stand and analyze the trade-offs between supervised and unsupervised learning, we examine its computational scalability, optimization dynamics, and conceptual sample complexity.

\subsection{Scalability Analysis}\label{sec:scalability}
The computational efficiency of \stand stems directly from its streamlined recurrent architecture, avoiding the quadratic complexity of standard self-attention mechanisms.

Let the input time series be denoted by $\mathbf{X} \in \mathbb{R}^{T \times C}$, where $T$ represents the temporal length and $C$ is the number of variables. We denote the hidden dimension of the model as $d$ and the number of layers in the Temporal Encoding Module as $L$. 

The complexity of the Feature Embedding module, which projects inputs via a multi-layer perceptron, is computed pointwise. For the entire sequence, the computational cost $\mathcal{C}_{\text{embed}}$ is formulated as:
\begin{equation}
    \mathcal{C}_{\text{embed}} = \mathcal{O}(T \cdot C \cdot d)
\end{equation}

Following the embedding, the bidirectional LSTM employed in the Temporal Encoding Module processes the sequence recurrently. The state update for a single LSTM cell involves matrix-vector multiplications proportional to the square of the hidden dimension. Consequently, for a sequence of length $T$ and a bidirectional configuration across $L$ layers, the temporal processing cost $\mathcal{C}_{\text{temp}}$ is bounded by:
\begin{equation}
    \mathcal{C}_{\text{temp}} = \mathcal{O}(T \cdot d^2 \cdot L)
\end{equation}

The Anomaly Scoring Module acts as a linear projection from the hidden space to the scalar anomaly score, contributing a cost $\mathcal{C}_{\text{score}}$:
\begin{equation}
    \mathcal{C}_{\text{score}} = \mathcal{O}(T \cdot d)
\end{equation}

By aggregating these components, the total time complexity per training epoch, denoted as $\mathcal{T}_{\text{total}}$, is derived as:
\begin{equation}
    \mathcal{T}_{\text{total}} = \mathcal{C}_{\text{embed}} + \mathcal{C}_{\text{temp}} + \mathcal{C}_{\text{score}} \approx \mathcal{O}(T(Cd + Ld^2))
\end{equation}

This computational complexity scales linearly with the sequence length $T$. While recurrent formulations inherently entail sequential step transitions along the time dimension ($h_t = f(h_{t-1}, x_t)$), preventing full temporal parallelization on GPU tensor cores, practical TSAD setups operate over compact sliding windows ($W \le 100$). As demonstrated empirically in Section \ref{sec:efficiency}, this bounded receptive field keeps sequential transition latency negligible (only $8.8\,\mu\text{s}$ per timestamp at $W=100$), while avoiding the quadratic attention bottleneck and multi-gigabyte memory footprint inherent in Transformer-based architectures ($\mathcal{T}_{\text{Trans}} \approx \mathcal{O}(T^2 \cdot d)$):
\begin{equation}
    \mathcal{T}_{\text{Trans}} \approx \mathcal{O}(T^2 \cdot d)
\end{equation}
Consequently, \stand circumvents the memory and latency bottlenecks of quadratic attention matrices for long input streams without incurring prohibitive recurrent delay.

Regarding memory constraints, the space complexity consists of parameter storage $\mathcal{M}_{\text{param}}$ and runtime activation memory $\mathcal{M}_{\text{run}}$. The parameter space is dominated by the LSTM weights:
\begin{equation}
    \mathcal{M}_{\text{param}} = \mathcal{O}(L \cdot d^2 + C \cdot d)
\end{equation}
Meanwhile, the runtime memory required for storing hidden states for backpropagation scales strictly linearly:
\begin{equation}
    \mathcal{M}_{\text{run}} = \mathcal{O}(T \cdot d \cdot L)
\end{equation}
Similar to the computational cost, this linear memory dependence offers a crucial scalability advantage over Transformer-based models, which typically require $\mathcal{O}(T^2)$ memory for attention matrices. This ensures that \stand~maintains a manageable memory footprint even when processing massive data streams, further cementing its suitability for large-scale industrial deployments.

\subsection{Optimization Dynamics and Convergence Analysis}\label{sec:convergence}
Supervised training provides stable gradient dynamics by directly anchoring parameter updates to verified binary anomaly boundaries. We formulate this parameter optimization as an Empirical Risk Minimization (ERM) problem over the training partition $\mathcal{D} = \{(\mathbf{X}_i, \mathbf{y}_i)\}_{i=1}^N$. The empirical loss objective $J(\theta)$ is governed by the step-wise Binary Cross-Entropy (BCE) function:
\begin{equation}
\small
    J(\theta) = - \frac{1}{N T} \sum_{i=1}^N \sum_{t=1}^T \left[ y_{i,t} \log(\sigma(s_{i,t})) + (1 - y_{i,t}) \log(1 - \sigma(s_{i,t})) \right]
\end{equation}
where $\theta$ represents the aggregated learnable weights, $s_{i,t}$ is the scalar logit, and $\sigma(\cdot)$ is the sigmoid operator. In our implementation, model parameters $\theta$ are trained using mini-batch stochastic backpropagation with the Adam optimizer \citep{adam}, which adaptively scales coordinate-wise learning rates based on running estimates of first and second gradient moments.

To characterize the underlying optimization stability under first-order gradient dynamics, consider standard gradient descent with learning rate $\eta$. Assuming the surrogate objective gradient $\nabla J(\theta)$ is $\mathcal{K}$-Lipschitz continuous ($\|\nabla J(\theta_1) - \nabla J(\theta_2) \| \leq \mathcal{K} \| \theta_1 - \theta_2 \|$ for $\mathcal{K} > 0$), invoking the classical Descent Lemma asserts an upper bound on successive parameter updates:
\begin{equation}
    J(\theta_{k+1}) \leq J(\theta_k) - \eta \left( 1 - \frac{\mathcal{K} \eta}{2} \right) \| \nabla J(\theta_k) \|^2
\end{equation}
Configuring the step size such that $\eta < \frac{2}{\mathcal{K}}$ guarantees that the loss trajectory $\{J(\theta_k)\}$ is monotonically non-increasing. Because the BCE loss is bounded below by zero, summing objective decrements across iterations guarantees convergence to a first-order stationary critical point:
\begin{equation}
\small
    \lim_{K \to \infty} \sum_{k=0}^{K} \| \nabla J(\theta_k) \|^2 \leq \frac{J(\theta_0)}{\eta ( 1 - \frac{\mathcal{K} \eta}{2} )} < \infty \implies \lim_{k \to \infty} \| \nabla J(\theta_k) \| = 0
    \label{eq:convergence}
\end{equation}
We clarify that Eq.~(\ref{eq:convergence}) reflects classical convergence properties of smooth first-order ERM objectives rather than an absolute theoretical proof of optimization superiority over unsupervised objectives (which similarly satisfy Lipschitz descent under mean squared error reconstruction). Rather than a theoretical rate disparity, the empirical benefit of supervisory feedback stems from its objective alignment: direct classification supervision penalizes misclassified operational transitions directly, avoiding the indirect surrogate proxy of minimizing input reconstruction errors where anomalous timestamps may blend into high-variance background dynamics.

\subsection{Sample Complexity and the Benefit of Supervision}\label{sec:sample_complexity}
To provide conceptual intuition regarding the empirical advantages observed when incorporating limited supervision, we examine the intrinsic learning difficulty of distribution modeling versus discriminative boundary learning. By comparing their respective sample complexities, we provide theoretical insight into why direct anomaly supervision can yield substantial performance dividends in high-dimensional or noisy environments.

Unsupervised methods that operate via statistical or density-based modeling fundamentally aim to capture the intrinsic distribution of normal data. For this class of methods, identifying outliers corresponds conceptually to non-parametric density estimation. According to minimax lower bounds established for nonparametric estimators \citep{g4}, estimating an arbitrary continuous density function in $\mathbb{R}^C$ with an expected error bound $\epsilon$ requires a sample size $N_{\text{UTAD}}$ satisfying:
\begin{equation}
N_{\text{UTAD}} = \Omega\left(\epsilon^{-\frac{C}{s}}\right)
\label{eq:utad_n}
\end{equation}
where $s$ denotes the smoothness parameter. This exponential dependence on the feature dimension $C$ illustrates the curse of dimensionality. While modern deep architectures employ inductive biases to mitigate this issue, mapping high-dimensional normal manifolds (e.g., $C=122$ in WADI) remains an inherently challenging sample-hungry task, often rendering purely unsupervised estimators sensitive to background noise and false positives.

In contrast, introducing supervisory signals reframes the learning objective into discriminative boundary identification. Under the agnostic Probably Approximately Correct (PAC) learning framework \citep{g3}, for a hypothesis class $\mathcal{H}$ with finite Vapnik-Chervonenkis (VC) dimension $d_{\text{VC}}(\mathcal{H})$, the labeled sample size $N_{\text{STAD}}$ required to ensure the generalization error is bounded by $\epsilon$ with confidence $1-\delta$ scales as:
\begin{equation}
N_{\text{STAD}} = \mathcal{O}\left(\frac{d_{\text{VC}}(\mathcal{H}) + \ln(1/\delta)}{\epsilon^2}\right)
\label{eq:stad_n}
\end{equation}
This PAC sample complexity bound depends polynomially on $1/\epsilon$ and linearly on model capacity $d_{\text{VC}}(\mathcal{H})$, avoiding direct exponential dependence on the input dimension $C$.

\textbf{Remark on Theoretical Intuition:} While Eq. (\ref{eq:utad_n}) and Eq. (\ref{eq:stad_n}) describe distinct learning objectives under different theoretical assumptions, their comparison offers valuable intuition into the observed empirical behavior:
\begin{enumerate}
    \item {Supervisory Return (Sec. \ref{sec:sup_gain}):} It provides intuition as to why our minimalist \stand baseline, even when trained on a restricted label budget (e.g., subset SP1 utilizing only 10\% of the data), can achieve competitive or superior performance compared to complex unsupervised architectures tasked with characterizing the entire unconstrained normal data distribution.
    \item {Noise Resilience (Sec. \ref{sec:noise_analysis}):} Because discriminative boundaries focus parameter capacity directly on the separating surface rather than full volume coverage, the learned boundary exhibits robust stability against moderate random label noise, as empirically validated under our noise perturbation settings.
\end{enumerate}

\section{Experiments}
\subsection{Experimental Objectives}\label{sec:exp_objectives}
To evaluate the performance characteristics of supervised and unsupervised paradigms, the empirical study is organized around four key objectives:

\begin{itemize}
\item \textbf{Benchmark Comparison:} Evaluating the overarching performance gap between supervised and unsupervised paradigms. STAD methods are trained on the preceding chronological training partition (50\% historical prefix), while UTAD methods operate under their standard label-free assumptions.
\item \textbf{Supervisory Gain Analysis:} Quantifying label efficiency by drastically reducing the available training data (e.g., utilizing as little as 10\%) on specifically partitioned subsets of the PSM dataset.
\item \textbf{Noise Robustness Evaluation:} Assessing the resilience of STAD methods against imperfect annotations by simulating realistic annotator fallibility via point and segment label noise.
\item \textbf{Model Diagnostics and Efficiency Profiling:} Validating the architectural design and deployment viability of \stand through component ablation, hyperparameter sensitivity, qualitative visualization, and empirical profiling of computational and memory overhead.
\end{itemize}

\subsection{Experimental Settings}

\subsubsection{Datasets}\label{sec:dataset}
To evaluate detection performance and scalability under operational conditions, we conduct evaluations across five multivariate time series benchmarks from the TSB-AD suite \citep{TSB-UAD}: PSM, SWaT, WADI, Swan, and Water. These datasets span diverse operational scales, with dimensionality ranging from 9 variables in Water to 122 variables in WADI. Each telemetry channel is standardized via z-score normalization ($\mu=0, \sigma=1$), where normalization parameters are estimated strictly on the training partition to prevent look-ahead leakage. Missing observations are imputed using forward-fill followed by backward-fill. To reflect continuous deployment and eliminate temporal look-ahead leakage, all evaluations are structured under a temporal out-of-sample protocol. Specifically, each telemetry stream is chronologically partitioned into an initial 50\% training prefix and a subsequent 50\% held-out evaluation suffix. Supervised models (STAD) are trained exclusively on the labeled prefix, while unsupervised baselines (UTAD-II) pre-train on the standard nominal sequence from TSB-AD; all models are evaluated on the unseen held-out suffix. Table~\ref{tab:dataset1} provides the structural dimensions, partition boundaries, point-level class proportions, and discrete anomaly episode counts across all five datasets.

\begin{table*}[htbp]
  \centering
  \caption{Characteristics, temporal out-of-sample partitioning, and discrete event counts for the five benchmark datasets. STAD models are trained on the initial 50\% temporal prefix and evaluated on the strictly held-out 50\% temporal suffix with zero temporal look-ahead leakage. UTAD-II models are pre-trained on the standard nominal sequence from TSB-AD \citep{TSB-UAD} and evaluated on the same held-out suffix.}
  \resizebox{\linewidth}{!}{
    \begin{tabular}{l|c|c|cccc|cccc|c}
    \toprule
    \multirow{2}{*}{\textbf{Dataset}} & \multirow{2}{*}{\textbf{\#Ch}} & \textbf{UTAD-II Train} & \multicolumn{4}{c|}{\textbf{Supervised Training Window (50\% Prefix)}} & \multicolumn{4}{c|}{\textbf{Held-out Evaluation Window (50\% Suffix)}} & \textbf{Total} \\
    \cmidrule(lr){4-7}\cmidrule(lr){8-11}
     & & \textbf{\#Timestamps} & \textbf{\#Steps} & \textbf{Normal (\%)} & \textbf{Anomaly (\%)} & \textbf{Events} & \textbf{\#Steps} & \textbf{Normal (\%)} & \textbf{Anomaly (\%)} & \textbf{Events} & \textbf{Events} \\
    \midrule
    PSM   & 25  & 132,481 & 43,920  & 32,991 (75.12\%) & 10,929 (24.88\%) & 43   & 43,921  & 30,469 (69.37\%) & 13,452 (30.63\%) & 29   & 72 \\
    SWaT  & 50  & 495,000 & 224,959 & 213,202 (94.77\%) & 11,757 (5.23\%)  & 21   & 224,960 & 182,096 (80.95\%) & 42,864 (19.05\%) & 14   & 35 \\
    WADI  & 122 & 241,921 & 17,280  & 15,944 (92.27\%)  & 1,336 (7.73\%)   & 7    & 17,281  & 16,634 (96.26\%)  & 647 (3.74\%)     & 6    & 13 \\
    Swan  & 38  & 60,000  & 30,000  & 25,178 (83.93\%)  & 4,822 (16.07\%)  & 3,696 & 30,000  & 15,262 (50.87\%)  & 14,738 (49.13\%) & 2,443 & 6,139 \\
    Water & 9   & 69,260  & 34,630  & 33,977 (98.11\%)  & 653 (1.89\%)     & 21   & 34,631  & 34,554 (99.78\%)  & 77 (0.22\%)      & 1    & 22 \\
    \bottomrule
    \end{tabular}%
  }
  \label{tab:dataset1}%
\end{table*}

To evaluate model behavior under acute label scarcity, we further construct a controlled scaling benchmark on PSM (Table~\ref{tab:dataset2}). In operational telemetry, adjacent anomalous timestamps typically represent continuous manifestations of an underlying physical breakdown rather than independent occurrences. Evaluating supervision purely by timestamp volume therefore risks overstating available supervisory information. To establish a realistic supervisory gradient, subsets SP1--SP3 incrementally scale the observed failure diversity from a sparse initial event budget to broader failure coverage, evaluating out-of-sample generalization across the remaining unobserved sequence.
\begin{table*}[htbp]
  \centering
  \caption{Characteristics and event-level statistics of the PSM subsets. Train Events and Rem. Events denote the count of independent, contiguous anomaly episodes in the training and remaining evaluation partitions, respectively. For SP2, one continuous anomaly episode spans the boundary at time step $20,182$, resulting in $23$ training events and $50$ remaining events from the $72$ overall continuous episodes in PSM.}
  \resizebox{\linewidth}{!}{
    \begin{tabular}{c|c|c|c|c|c|c|c|c|c|c}
    \toprule
    \textbf{Dataset} & \textbf{Split Threshold} & \textbf{\#Train} & \textbf{Train Anomaly Rate} & \textbf{Train Events} & \textbf{\#Remaining} & \textbf{Rem. Anomaly Rate} & \textbf{Rem. Events} & \textbf{\#Overall} & \textbf{Total Events} & \textbf{Overall Anomaly Rate} \\
    \midrule
    SP1   & 10.00\% & 8784  & 13.35\% & 12 & 79057 & 29.36\% & 60 & \multirow{3}{*}{87841} & \multirow{3}{*}{72} & \multirow{3}{*}{27.76\%} \\
    SP2   & 20.00\% & 20182 & 20.00\% & 23 & 67659 & 30.07\% & 50 & & & \\
    SP3   & 30.00\% & 35136 & 30.73\% & 35 & 52705 & 25.77\% & 37 & & & \\
    \bottomrule
    \end{tabular}}
  \label{tab:dataset2}%
\end{table*}%

\subsubsection{Baselines}\label{app:baseline}

The UTAD-I methods consist of {traditional unsupervised anomaly detection approaches}, covering Random Guess (Random), a method that predicts each time point as anomalous with a 50\% probability and serves as an intuitive baseline to contrast the performance gaps between other methods and a random strategy, Isolation Forest (IForest), Local Outlier Factor (LOF), Principal Component Analysis (PCA), Histogram-based Outlier Score (HBOS), K-Nearest Neighbors (KNN), and K-Means Clustering (KMeans). These methods rely on classical statistical characteristics, distance metrics, or clustering mechanisms to identify anomalies without requiring labeled training data.

The UTAD-II methods encompass a broader range of techniques, including {traditional kernel methods, deep learning models, and Transformer-based approaches}, specifically including One-Class Support Vector Machine (OCSVM), Autoencoder (AE), Convolutional Neural Network (CNN), Long Short-Term Memory Network (LSTM), TranAD \citep{trans_ad}, USAD \citep{usad}, Omni \citep{omni}, Anomaly Transformer (A.T.) \citep{anom_trans}, TimesNet \citep{timesnet}, M2N2 \citep{M2N2}, LFTSAD \citep{LFTSAD}, and CATCH \citep{catch}. This category integrates shallow kernel-based learning, deep neural networks (for spatial-temporal feature extraction), and state-of-the-art Transformer architectures (for capturing long-range dependencies in time series).

The STAD methods are composed of {traditional supervised learning classification models}, including Random Forest (RF), Support Vector Machine (SVM), AdaBoost, Extra-Trees Classifier (ExtraTrees), and Histogram-based Gradient Boosting Classification Tree (LightGBM). These models are trained on labeled normal/anomalous samples to learn discriminative patterns for direct anomaly classification.

\textbf{Implementation Details:}
\begin{itemize}
    \item All methods in UTAD-I, as well as OCSVM, AE, CNN, LSTM, TranAD, USAD, Omni, Anomaly Transformer (A.T.), and TimesNet in UTAD-II, adopt the implementations provided by the TSB-AD benchmark library\footnote{\url{https://github.com/TheDatumOrg/TSB-AD}}. TSB-AD is a widely recognized time-series anomaly detection benchmark that ensures consistent experimental settings (\textit{e.g.}, hyperparameter ranges, data preprocessing) across different models, enhancing the reproducibility of our comparisons.
    \item For M2N2, LFTSAD, and CATCH in UTAD-II, we implement the models based on the official open-source code released by the authors of their original papers. We adhere to the recommended hyperparameter configurations, network architectures, and training protocols reported in the literature to maintain the authenticity of the baseline performances.
    \item All methods in STAD are implemented based on the Scikit-learn library\footnote{\url{https://scikit-learn.org/}}, a mature and extensively validated machine learning toolkit in the Python ecosystem. The Scikit-learn implementations are chosen for their stability, efficiency, and widespread adoption in academic and industrial research, ensuring reliable baseline results.
\end{itemize}

\subsubsection{Evaluation Metrics}\label{app:metric}
To comprehensively assess model performance across multiple granularities, we evaluate pointwise detection accuracy using standard F1 and AUC-ROC metrics; following established TSAD benchmark conventions \citep{TSB-UAD}, threshold-dependent F1 is reported under post-hoc optimal threshold search (Best-F1) to facilitate standardized comparability across disparate baseline models. For event-level localization, we utilize Aff-F1 \citep{aff}, its unbiased precision-focused refinement UAff-F1 \citep{simad}, and VUS-PR \citep{VUS} to effectively capture contiguous anomaly segments. Finally, we adopt CCE \citep{cce} to quantify prediction stability. By accounting for both local and global uncertainties, CCE ensures a robust evaluation particularly suited for real-world scenarios with noisy labels.

\subsection{Comparison of Unsupervised and Supervised Methods}\label{sec:comparison}
Table \ref{tab:comparison} presents the performance of representative UTAD-I, UTAD-II, and STAD methods evaluated on the held-out test horizon. Unsupervised methods rarely attain top performance across the benchmark datasets and frequently exhibit pronounced deficits on specific metric families. For instance, recent architectures such as M2N2 and CATCH register low CCE scores across all five datasets, while methods such as Omni and A.T. obtain deficient Aff-F1 and UAff-F1 scores on PSM, SWaT, and Swan.

Figure \ref{fig:comparison}(a) displays the average scores across datasets and metrics for all 20 evaluated models, along with category averages. Although UTAD-II methods leverage normal-state inductive priors and improve over UTAD-I (group mean of $39.30$ versus $32.09$), they remain less effective overall than standard linear projections such as PCA ($44.34$). In contrast, supervised methods deliver marked performance gains, yielding a group mean of $43.69$. Our minimalist baseline, \stand, attains the highest overall score of $54.68$, outperforming the UTAD-I average by $70.4\%$ ($+22.59$) and the UTAD-II average by $39.1\%$ ($+15.38$). Even against the single highest-performing unsupervised model, LSTM ($47.10$), \stand achieves an advantage of $7.58$ points.

Figure \ref{fig:comparison}(b) illustrates the multi-metric capability profile of top-performing representative methods from each category. On prediction consistency, \stand attains a CCE of $38.50$, whereas every unsupervised method scores below $17.55$, reflecting low prediction certainty and noisy thresholding in unsupervised detectors. Furthermore, on unbiased event-level localization, \stand achieves an UAff-F1 of $58.22$, which substantially surpasses LSTM ($46.09$), CATCH ($27.88$), and M2N2 ($24.80$).

To guarantee strict out-of-sample temporal evaluation without data leakage, all models are evaluated on an unobserved out-of-sample test horizon. For supervised methods, training utilizes a preceding labeled temporal window, while unsupervised and semi-supervised baselines strictly adhere to the standard training and inference protocols established by the TSB-UAD benchmark \citep{TSB-UAD}. This unified evaluation setting guarantees zero temporal data leakage. For readers seeking direct comparability with prior studies evaluated across the unbroken test sequence, full-sequence results are provided in Supplementary Section~\ref{app:split_validation}. Across both evaluation settings, \stand achieves superior overall average performance compared to unsupervised baselines and remains competitive with classical supervised alternatives.

To provide a comprehensive assessment and mitigate metric-specific evaluation biases, detection performance is examined across five distinct benchmark datasets and six complementary metrics. In addition, to assess run-to-run variability resulting from random initializations, all stochastic baselines and deep neural architectures are evaluated across $N=5$ independent runs with distinct random seeds ($\{40, 41, 42, 43, 44\}$). Table \ref{tab:comparison} reports the resulting mean performance across all evaluated settings, while complete dataset-level and cross-dataset standard deviations across independent runs alongside standardized effect sizes (Cohen's $d$) are documented in Supplementary Table~\ref{tab:left_test_std} and Supplementary Section~\ref{app:effect_size}. Furthermore, to complement aggregate score averages with a scale-invariant assessment, we conduct a non-parametric ranking analysis across all 30 dataset-metric evaluation settings, as detailed in Supplementary Table \ref{tab:ranking_summary}. Under this ranking protocol, \stand achieves the top overall average rank of $5.70$ across all 20 evaluated models, surpassing the leading unsupervised baseline at $7.77$ and confirming that its empirical advantage is robust across disparate metric scales.

Finally, \stand demonstrates notable advantages on SWaT and WADI, which feature larger sample volumes, while tree ensembles such as ExtraTrees and Random Forest remain competitive on smaller datasets. This pattern indicates that while classical supervised ensembles provide dependable baselines in low-sample regimes, the representation capacity of deep supervised modeling in \stand translates larger observation volumes into substantial performance gains. In addition, on Water, which presents extreme class imbalance ($0.22\%$ anomaly ratio in the evaluation split), \stand attains a CCE of $64.19$ alongside an Aff-F1 of $92.21$, indicating stable nominal predictions across the $34,554$ normal time steps while isolating the solitary failure episode.
\begin{table*}[htbp]
  \centering
  \caption{Performance comparison of the three method categories across five datasets and six metrics evaluated on the held-out test horizon. \textbf{Bold} indicates the best performance, and \uline{underlining} indicates the second-best performance. Methods highlighted in gray are UTAD-I, methods highlighted in green are UTAD-II, and methods highlighted in orange represent STAD methods.}
  \resizebox{\linewidth}{!}{
    \begin{tabular}{c|cccccc|cccccc|cccccc}
    \toprule
    \textbf{Dataset} & \multicolumn{6}{c|}{\textbf{PSM}}             & \multicolumn{6}{c|}{\textbf{SWaT}}            & \multicolumn{6}{c}{\textbf{WADI}} \\
    \midrule
    \textbf{Metric} & \textbf{CCE} & \textbf{F1} & \textbf{Aff-F1} & \textbf{UAff-F1} & \textbf{AUC} & \textbf{V-PR} & \textbf{CCE} & \textbf{F1} & \textbf{Aff-F1} & \textbf{UAff-F1} & \textbf{AUC} & \textbf{V-PR} & \textbf{CCE} & \textbf{F1} & \textbf{Aff-F1} & \textbf{UAff-F1} & \textbf{AUC} & \textbf{V-PR} \\
    \midrule
    \rowcolor[rgb]{ .949,  .949,  .949} Random    & -0.52  & 8.49  & 71.43  & 11.66  & 50.20  & 35.30  & -0.46  & 7.87  & 69.35  & -4.59  & 50.06  & 19.55  & 0.99  & 3.18  & 66.37  & 20.87  & 51.02  & 6.17  \\
    \rowcolor[rgb]{ .949,  .949,  .949} IForest   & 0.40  & 14.23  & 49.22  & -1.25  & 51.34  & 36.30  & 0.40  & 6.72  & 67.58  & -3.93  & 32.17  & 15.36  & 11.57  & 10.59  & 72.41  & 26.20  & 85.22  & 19.86  \\
    \rowcolor[rgb]{ .949,  .949,  .949} PCA       & 10.22  & 17.77  & 56.73  & 51.31  & 78.32  & 59.91  & 5.64  & 34.75  & 61.50  & -2.99  & 92.11  & 77.11  & 16.32  & 32.83  & 83.96  & 58.69  & 84.93  & 29.05  \\
    \rowcolor[rgb]{ .949,  .949,  .949} HBOS      & -0.50  & 10.66  & 54.69  & 8.80  & 48.40  & 34.71  & 14.58  & 5.07  & 71.07  & 26.69  & 85.23  & 51.92  & 16.01  & 18.53  & 75.88  & 35.34  & 85.02  & 26.56  \\
    \rowcolor[rgb]{ .949,  .949,  .949} KMeans    & -1.10  & 6.79  & 63.80  & 53.36  & 35.30  & 28.84  & 0.53  & 14.31  & \textbf{77.65}  & 40.88  & 21.37  & 17.78  & 13.50  & 5.43  & 60.69  & 0.36  & 73.85  & 10.92  \\
    \midrule
    \midrule
    \rowcolor[rgb]{ .886,  .937,  .855} AE        & 5.94  & 4.90  & 35.00  & 14.38  & 59.18  & 37.57  & 21.36  & 31.37  & 76.01  & 46.24  & 91.72  & 76.31  & 22.08  & 24.35  & 73.78  & 45.87  & 77.90  & 18.94  \\
    \rowcolor[rgb]{ .886,  .937,  .855} CNN       & 2.06  & \uline{24.97}  & 49.02  & 28.51  & 78.98  & \textbf{62.25}  & 37.30  & \uline{41.57}  & 11.57  & 11.57  & 91.75  & \textbf{82.35}  & 13.53  & 28.86  & \uline{86.69}  & 64.58  & \uline{87.09}  & 35.37  \\
    \rowcolor[rgb]{ .886,  .937,  .855} LSTM      & 1.87  & \textbf{25.54}  & 53.94  & 37.36  & 77.35  & \uline{61.26}  & 38.07  & \uline{41.57}  & 12.29  & 12.29  & 89.88  & \uline{81.90}  & 14.65  & 29.25  & 86.66  & 66.06  & 83.47  & 35.49  \\
    \rowcolor[rgb]{ .886,  .937,  .855} TranAD    & 0.47  & 16.40  & 47.32  & 43.45  & 69.66  & 53.91  & 38.32  & \uline{41.57}  & 12.26  & 12.26  & \uline{92.69}  & 81.38  & 12.60  & 27.93  & 85.20  & 66.75  & 84.60  & 36.17  \\
    \rowcolor[rgb]{ .886,  .937,  .855} USAD      & 4.28  & 16.72  & 44.31  & 40.91  & 72.39  & 55.40  & 38.30  & \uline{41.57}  & 12.23  & 12.23  & 92.40  & 80.85  & 15.29  & 30.31  & 86.20  & \uline{67.15}  & 76.46  & 29.52  \\
    \rowcolor[rgb]{ .886,  .937,  .855} Omni      & 11.96  & 18.16  & 17.55  & 16.05  & 72.61  & 51.57  & 38.41  & \uline{41.57}  & 11.16  & 11.16  & \textbf{92.78}  & 77.46  & 14.66  & 30.18  & 79.33  & 46.35  & 78.19  & 42.14  \\
    \rowcolor[rgb]{ .886,  .937,  .855} A.T.      & 0.51  & 10.26  & 63.57  & -3.85  & 50.65  & 36.48  & 0.00  & 1.20  & 65.24  & -14.25  & 49.26  & 19.53  & -0.34  & 1.06  & 35.84  & -53.36  & 36.60  & 5.76  \\
    \rowcolor[rgb]{ .886,  .937,  .855} M2N2      & -0.01  & 14.83  & \uline{80.84}  & 46.18  & 65.93  & 51.71  & -0.51  & 15.32  & 27.56  & 25.80  & 80.08  & 40.06  & 10.63  & 27.13  & 81.07  & 38.19  & 83.14  & 35.44  \\
    \rowcolor[rgb]{ .886,  .937,  .855} CATCH     & 0.54  & 15.26  & \textbf{83.83}  & \textbf{56.61}  & 61.46  & 50.61  & 0.01  & 5.79  & 72.34  & 10.56  & 19.74  & 13.57  & 2.23  & 15.75  & 76.92  & 39.90  & 80.45  & 26.05  \\
    \midrule
    \midrule
    \rowcolor[rgb]{ .988,  .894,  .839} RF        & \uline{13.14}  & 18.81  & 50.78  & 21.89  & 73.17  & 50.54  & 31.01  & 39.04  & 60.07  & \uline{56.87}  & 90.84  & 80.28  & 23.82  & \uline{46.73}  & 73.04  & 60.38  & 85.49  & \uline{63.63}  \\
    \rowcolor[rgb]{ .988,  .894,  .839} SVM       & 12.25  & 22.13  & 65.87  & 52.31  & \textbf{80.42}  & 57.39  & 19.93  & 15.09  & 68.89  & 25.96  & 91.67  & 76.17  & 12.87  & 33.22  & 81.05  & 50.59  & 81.50  & 42.92  \\
    \rowcolor[rgb]{ .988,  .894,  .839} AdaBoost  & 10.05  & 20.06  & 34.14  & 28.86  & 70.80  & 39.94  & 15.39  & 22.34  & 47.91  & 34.16  & 84.07  & 37.79  & 11.60  & 27.13  & 58.41  & 24.08  & 78.62  & 27.07  \\
    \rowcolor[rgb]{ .988,  .894,  .839} ExtraTrees& 10.78  & 20.36  & 53.19  & 36.77  & \uline{79.11}  & 57.64  & 39.10  & \textbf{52.71}  & 12.91  & 12.89  & 88.88  & 53.93  & \textbf{27.43}  & \textbf{51.47}  & \textbf{88.03}  & \textbf{70.48}  & \textbf{94.61}  & \textbf{73.56}  \\
    \rowcolor[rgb]{ .988,  .894,  .839} LightGBM  & 6.79  & 16.35  & 55.07  & 16.70  & 71.52  & 51.89  & \uline{39.25}  & 28.39  & 30.33  & -12.15  & 89.38  & 64.47  & 23.50  & 44.47  & 77.52  & 50.19  & 86.81  & 50.02  \\
    \rowcolor[rgb]{ .988,  .894,  .839} STAND     & \textbf{16.39}  & 23.58  & 64.18  & \uline{55.61}  & 66.81  & 55.47  & \textbf{51.97}  & 36.88  & \uline{76.94}  & \textbf{69.73}  & 90.68  & 79.62  & \uline{26.70}  & 37.33  & 78.10  & 55.51  & 73.17  & 39.71  \\
    \midrule
    \textbf{Dataset} & \multicolumn{6}{c|}{\textbf{Swan}}            & \multicolumn{6}{c|}{\textbf{Water}}           & \multicolumn{6}{c}{\textbf{Average}} \\
    \midrule
    \textbf{Metric} & \textbf{CCE} & \textbf{F1} & \textbf{Aff-F1} & \textbf{UAff-F1} & \textbf{AUC} & \textbf{V-PR} & \textbf{CCE} & \textbf{F1} & \textbf{Aff-F1} & \textbf{UAff-F1} & \textbf{AUC} & \textbf{V-PR} & \textbf{CCE} & \textbf{F1} & \textbf{Aff-F1} & \textbf{UAff-F1} & \textbf{AUC} & \textbf{V-PR} \\
    \midrule
    \rowcolor[rgb]{ .949,  .949,  .949} Random    & -0.10  & 9.10  & 26.79  & 4.24  & 49.71  & 89.09  & 4.74  & 0.55  & 66.75  & 0.38  & 55.06  & 0.49  & 0.93  & 5.84  & 60.14  & 6.51  & 51.21  & 30.12  \\
    \rowcolor[rgb]{ .949,  .949,  .949} IForest   & 5.77  & 17.29  & 3.23  & -3.03  & 68.63  & 92.47  & 37.09  & 8.07  & 52.82  & -47.52  & 98.97  & 17.21  & 11.05  & 11.38  & 49.05  & -5.91  & 67.27  & 36.24  \\
    \rowcolor[rgb]{ .949,  .949,  .949} PCA       & 1.52  & 18.20  & 1.35  & 1.33  & 53.33  & 96.02  & 46.62  & 7.74  & 77.31  & 41.62  & \uline{99.01}  & 37.89  & 16.06  & 22.26  & 56.17  & 29.99  & 81.54  & 60.00  \\
    \rowcolor[rgb]{ .949,  .949,  .949} HBOS      & 13.70  & 15.04  & 18.22  & 13.47  & 80.02  & 96.18  & 31.77  & 6.85  & 56.58  & -39.18  & 96.26  & 9.63  & 15.11  & 11.23  & 55.29  & 9.02  & 78.99  & 43.80  \\
    \rowcolor[rgb]{ .949,  .949,  .949} KMeans    & -9.32  & 4.74  & 16.58  & 8.88  & 23.50  & 87.55  & 44.24  & 6.19  & 39.83  & -66.82  & 93.74  & 24.05  & 9.57  & 7.49  & 51.71  & 7.33  & 49.55  & 33.83  \\
    \midrule
    \midrule
    \rowcolor[rgb]{ .886,  .937,  .855} AE        & 13.97  & 13.59  & 21.21  & 20.46  & 68.29  & 92.43  & 17.22  & 1.88  & 71.53  & 22.88  & 58.07  & 2.88  & 16.11  & 15.22  & 55.51  & 29.97  & 71.03  & 45.63  \\
    \rowcolor[rgb]{ .886,  .937,  .855} CNN       & 0.69  & 18.42  & 0.08  & 0.08  & 66.83  & 96.12  & 31.41  & 8.51  & 90.19  & 78.80  & \textbf{99.87}  & \textbf{59.99}  & 17.00  & \textbf{24.47}  & 47.51  & 36.71  & \textbf{84.90}  & \textbf{67.22}  \\
    \rowcolor[rgb]{ .886,  .937,  .855} LSTM      & 0.50  & 16.37  & 29.49  & \uline{29.11}  & 71.67  & 95.29  & 27.28  & 5.31  & 93.18  & 85.62  & 91.41  & 18.73  & 16.47  & 23.61  & 55.11  & \uline{46.09}  & 82.76  & 58.53  \\
    \rowcolor[rgb]{ .886,  .937,  .855} TranAD    & 0.08  & 11.20  & 19.84  & -7.58  & 60.35  & 94.03  & 19.12  & 5.20  & 91.42  & 81.54  & 94.48  & 20.85  & 14.12  & 20.46  & 51.21  & 39.28  & 80.35  & 57.27  \\
    \rowcolor[rgb]{ .886,  .937,  .855} USAD      & 0.70  & 16.44  & 3.97  & 3.81  & 56.80  & 95.05  & 29.06  & 5.53  & 90.52  & 79.71  & 93.95  & 20.92  & 17.53  & 22.11  & 47.45  & 40.76  & 78.40  & 56.35  \\
    \rowcolor[rgb]{ .886,  .937,  .855} Omni      & 2.45  & 18.34  & 0.08  & 0.08  & 59.54  & 95.63  & 18.24  & 3.98  & 92.22  & 83.31  & 92.90  & 42.93  & 17.14  & 22.45  & 40.07  & 31.39  & 79.20  & \uline{61.95}  \\
    \rowcolor[rgb]{ .886,  .937,  .855} A.T.      & 0.01  & 5.64  & 18.16  & -12.74  & 49.19  & 89.65  & 15.94  & 2.99  & 51.39  & -43.74  & 61.45  & 2.99  & 3.22  & 4.23  & 46.84  & -25.59  & 49.43  & 30.88  \\
    \rowcolor[rgb]{ .886,  .937,  .855} M2N2      & 0.53  & 18.45  & 0.08  & 0.08  & 85.05  & \textbf{98.41}  & 7.69  & 5.64  & 70.36  & 13.77  & 91.45  & 22.53  & 3.67  & 16.27  & 51.98  & 24.80  & 81.13  & 49.63  \\
    \rowcolor[rgb]{ .886,  .937,  .855} CATCH     & 0.40  & \uline{18.48}  & 0.08  & 0.08  & 55.82  & 95.67  & 22.06  & 5.53  & 75.30  & 32.23  & 89.20  & \uline{44.83}  & 5.05  & 12.16  & \uline{61.69}  & 27.88  & 61.33  & 46.15  \\
    \midrule
    \midrule
    \rowcolor[rgb]{ .988,  .894,  .839} RF        & 27.51  & 16.68  & \uline{29.91}  & 28.55  & \uline{87.60}  & 95.32  & \uline{53.80}  & \uline{19.66}  & 65.81  & 60.20  & 87.16  & 26.33  & \uline{26.14}  & 17.93  & 54.32  & 40.44  & 84.07  & 51.85  \\
    \rowcolor[rgb]{ .988,  .894,  .839} SVM       & 12.31  & 16.55  & 25.65  & 24.73  & 82.46  & \uline{97.19}  & 19.17  & 7.24  & 66.59  & 42.27  & 65.06  & 12.07  & 11.31  & 12.00  & 51.72  & 33.02  & 77.30  & 47.20  \\
    \rowcolor[rgb]{ .988,  .894,  .839} AdaBoost  & 33.24  & \textbf{20.19}  & 27.23  & 25.98  & 87.31  & 93.06  & 33.52  & 17.43  & \textbf{98.42}  & \textbf{96.84}  & 88.12  & 32.65  & 17.59  & 14.29  & 46.72  & 28.74  & 78.36  & 45.24  \\
    \rowcolor[rgb]{ .988,  .894,  .839} ExtraTrees& 27.90  & 17.86  & 28.26  & 26.99  & 87.35  & 95.54  & 26.57  & 13.18  & \uline{97.91}  & \uline{96.04}  & 85.31  & 11.49  & 25.52  & 18.39  & 50.63  & 35.95  & \uline{84.59}  & 52.82  \\
    \rowcolor[rgb]{ .988,  .894,  .839} LightGBM  & \textbf{38.09}  & 16.29  & \textbf{36.29}  & \textbf{35.21}  & \textbf{89.54}  & 94.90  & 21.21  & \textbf{26.39}  & 33.03  & -2.69  & 80.66  & 3.80  & 24.28  & 14.79  & 49.23  & 21.57  & 80.89  & 47.88  \\
    \rowcolor[rgb]{ .988,  .894,  .839} STAND     & \uline{33.26}  & 17.17  & 27.39  & 26.95  & 77.27  & 96.69  & \textbf{64.19}  & 6.63  & 92.21  & 83.33  & 96.53  & 20.37  & \textbf{38.50}  & \uline{24.32}  & \textbf{67.76}  & \textbf{58.22}  & 80.89  & 58.37  \\
    \bottomrule
    \end{tabular}
    }
  \label{tab:comparison}
\end{table*}

\subsection{Supervisory Scaling and Few-Shot Label Efficiency}\label{sec:sup_gain}
In real-world telemetry archives, true failure events are historically sparse; many operational anomalies that occurred in past years are never archived or documented in maintenance logs. To examine whether supervised models remain viable when historical archives provide only a constrained budget of failure incidents, we conduct controlled scaling evaluations on PSM and SWaT (Table~\ref{tab:dataset2}, Table~\ref{tab:sp3_comparison}, and Section~\ref{sec:fewshot_efficiency}). Here, labels within the training partition are clean, evaluating model generalization across unobserved future failure events under restricted event coverage.

As shown in Table \ref{tab:sp3_comparison}, the incorporation of supervisory signals yields substantial performance improvements for STAD models under temporal adaptation. We can draw several key observations from these results. 

First, as supervisory signals expand from SP1 to SP3, increasing from $12$ training events across $8,784$ time steps to $23$ events in SP2 and $35$ events in SP3, the performance of supervised methods demonstrates a clear and consistent upward trajectory. Specifically, in SP1, \stand observes only $12$ discrete failure episodes while evaluated on $60$ unobserved episodes, representing an acute unobserved-event scenario where $83.3\%$ of failure episodes remain completely unrecorded in the training archive. Even under this constrained event budget, \stand attains an average score of $39.94$ and an average rank of $3.0$, outperforming deep unsupervised architectures on remaining unseen events (such as CATCH at an average rank of $3.8$). As the number of observed failure episodes expands to $35$ in SP3, \stand's average score increases to $47.89$, maintaining the top average rank of $2.5$. This demonstrates that supervisory returns scale systematically with both temporal sequence length and independent failure episode diversity.

Second, the results reveal a distinct divergence among unsupervised methods based on data volume. When the training sequence is restricted, as in SP1, traditional UTAD-I methods such as PCA perform remarkably well, achieving a competitive average rank of $2.2$. In contrast, deep learning-based UTAD-II methods, such as CATCH, require substantial historical observations to reconstruct normal operational manifolds effectively. Consequently, their performance is constrained in low-data regimes, and even as sequence length increases in SP2 and SP3, their overall performance still falls short of supervised alternatives.

Finally, unsupervised methods such as CATCH and PCA exhibit stagnant or fluctuating performance across the subsets because they cannot directly optimize classification boundaries using anomaly feedback. In contrast, STAD methods consistently capitalize on the expanded label set. In both SP2 and SP3, \stand achieves leading average scores of $46.67$ and $47.89$, alongside the top average ranks of $1.8$ and $2.5$, respectively. This empirical evidence confirms that when limited anomaly labels are accessible, utilizing sparse event supervision delivers substantial practical returns, often proving more cost-effective than escalating unsupervised model complexity.

\begin{table}[htbp]
  \centering
  \caption{Performance comparison across the three PSM subsets (SP1--SP3). \textbf{Bold} indicates the best performance, and \uline{underlining} indicates the second-best performance. Methods highlighted in gray are UTAD-I, green are UTAD-II, and plain are STAD. Avg.~Score denotes the unweighted arithmetic mean across all six evaluation metrics ($1/6 \sum_{m=1}^6 \text{Score}_m$), while Avg.~Rk. denotes the non-parametric average rank across the six metrics within each partition (lower rank indicates superior performance).}
  \resizebox{\linewidth}{!}{
\begin{tabular}{c|c|cccccc|c|c}
\toprule
\textbf{Source} & \multicolumn{1}{c}{\textbf{Model}} & \textbf{CCE} & \textbf{F1} & \textbf{Aff-F1} & \textbf{UAff-F1} & \textbf{AUC-ROC} & \textbf{VUS-PR} & \textbf{Avg. Score} & \textbf{Avg. Rk.} \\
\midrule
\multirow{8}[5]{*}{SP1} & \cellcolor[rgb]{ .949,  .949,  .949}PCA & \cellcolor[rgb]{ .949,  .949,  .949}5.47  & \cellcolor[rgb]{ .949,  .949,  .949}\uline{18.17} & \cellcolor[rgb]{ .949,  .949,  .949}\uline{58.88} & \cellcolor[rgb]{ .949,  .949,  .949}\uline{53.51} & \cellcolor[rgb]{ .949,  .949,  .949}\textbf{73.35} & \cellcolor[rgb]{ .949,  .949,  .949}\textbf{55.78} & \cellcolor[rgb]{ .949,  .949,  .949}\uline{44.19} & \cellcolor[rgb]{ .949,  .949,  .949}\textbf{2.2} \\
\cmidrule{2-10}      & \cellcolor[rgb]{ .886,  .937,  .855}CATCH & \cellcolor[rgb]{ .886,  .937,  .855}0.59  & \cellcolor[rgb]{ .886,  .937,  .855}16.89  & \cellcolor[rgb]{ .886,  .937,  .855}\textbf{82.45} & \cellcolor[rgb]{ .886,  .937,  .855}\textbf{55.80} & \cellcolor[rgb]{ .886,  .937,  .855}62.88  & \cellcolor[rgb]{ .886,  .937,  .855}52.47  & \cellcolor[rgb]{ .886,  .937,  .855}\textbf{45.18} & \cellcolor[rgb]{ .886,  .937,  .855}3.8 \\
\cmidrule{2-10}      & RF    & 4.64  & 9.89  & 46.56  & 34.29  & 60.37  & 41.24  & 32.83  & 6.5 \\
      & SVM   & 6.09  & 11.69  & 42.43  & 31.52  & 63.88  & 45.37  & 33.50  & 5.8 \\
      & AdaBoost & \uline{6.76} & 15.61  & 38.38  & 31.95  & \uline{71.18} & 49.58  & 35.58  & 4.2 \\
      & ExtraTrees & 6.33  & 16.26  & 46.13  & 30.89  & 66.19  & 47.01  & 35.47  & 4.8 \\
      & LightGBM & 4.71  & 10.15  & 56.95  & 17.20  & 66.20  & 45.81  & 33.50  & 5.7 \\
      & STAND & \textbf{14.83} & \textbf{19.68} & 54.18  & 29.74  & 68.17  & \uline{53.01} & 39.94  & \uline{3.0} \\
\midrule
\multirow{8}[6]{*}{SP2} & \cellcolor[rgb]{ .949,  .949,  .949}PCA & \cellcolor[rgb]{ .949,  .949,  .949}5.51  & \cellcolor[rgb]{ .949,  .949,  .949}18.71  & \cellcolor[rgb]{ .949,  .949,  .949}\uline{58.32} & \cellcolor[rgb]{ .949,  .949,  .949}\uline{52.88} & \cellcolor[rgb]{ .949,  .949,  .949}74.29  & \cellcolor[rgb]{ .949,  .949,  .949}57.11  & \cellcolor[rgb]{ .949,  .949,  .949}44.47  & \cellcolor[rgb]{ .949,  .949,  .949}3.8 \\
\cmidrule{2-10}      & \cellcolor[rgb]{ .886,  .937,  .855}CATCH & \cellcolor[rgb]{ .886,  .937,  .855}0.97  & \cellcolor[rgb]{ .886,  .937,  .855}17.34  & \cellcolor[rgb]{ .886,  .937,  .855}\textbf{83.75} & \cellcolor[rgb]{ .886,  .937,  .855}\textbf{59.94} & \cellcolor[rgb]{ .886,  .937,  .855}63.21  & \cellcolor[rgb]{ .886,  .937,  .855}54.86  & \cellcolor[rgb]{ .886,  .937,  .855}\textbf{46.68} & \cellcolor[rgb]{ .886,  .937,  .855}4.8 \\
\cmidrule{2-10}      & RF    & 10.27  & 17.00  & 47.99  & 22.77  & 71.47  & 51.47  & 36.83  & 6.0 \\
      & SVM   & 10.21  & 17.19  & 44.59  & 31.41  & 76.14  & 55.27  & 39.13  & 5.5 \\
      & AdaBoost & 7.85  & 16.91  & 28.97  & 22.64  & 69.73  & 46.39  & 32.08  & 7.5 \\
      & ExtraTrees & 11.19  & \textbf{23.12} & 49.00  & 39.60  & \uline{77.26} & \uline{59.98} & 43.36  & \uline{2.7} \\
      & LightGBM & \uline{13.16} & 21.12  & 46.73  & 35.69  & 76.53  & 57.11  & 41.72  & 3.8 \\
      & STAND & \textbf{19.76} & \uline{21.81} & 50.26  & 42.67  & \textbf{80.48} & \textbf{65.01} & \uline{46.67} & \textbf{1.8} \\
\midrule
\multirow{8}[6]{*}{SP3} & \cellcolor[rgb]{ .949,  .949,  .949}PCA & \cellcolor[rgb]{ .949,  .949,  .949}11.09  & \cellcolor[rgb]{ .949,  .949,  .949}22.37  & \cellcolor[rgb]{ .949,  .949,  .949}58.27  & \cellcolor[rgb]{ .949,  .949,  .949}51.73  & \cellcolor[rgb]{ .949,  .949,  .949}79.03  & \cellcolor[rgb]{ .949,  .949,  .949}55.28  & \cellcolor[rgb]{ .949,  .949,  .949}46.29  & \cellcolor[rgb]{ .949,  .949,  .949}3.8 \\
\cmidrule{2-10}      & \cellcolor[rgb]{ .886,  .937,  .855}CATCH & \cellcolor[rgb]{ .886,  .937,  .855}2.60  & \cellcolor[rgb]{ .886,  .937,  .855}15.73  & \cellcolor[rgb]{ .886,  .937,  .855}\textbf{82.18} & \cellcolor[rgb]{ .886,  .937,  .855}\textbf{53.20} & \cellcolor[rgb]{ .886,  .937,  .855}60.61  & \cellcolor[rgb]{ .886,  .937,  .855}45.90  & \cellcolor[rgb]{ .886,  .937,  .855}43.37  & \cellcolor[rgb]{ .886,  .937,  .855}5.5 \\
\cmidrule{2-10}      & RF    & \uline{12.75} & 15.93  & 47.02  & 15.97  & 72.74  & 50.62  & 35.84  & 5.7 \\
      & SVM   & 12.25  & 22.13  & \uline{65.87} & \uline{52.09} & \uline{80.42} & 57.39  & \textbf{48.36} & \uline{2.7} \\
      & AdaBoost & 10.77  & 17.95  & 38.86  & 27.34  & 71.87  & 43.73  & 35.09  & 7.0 \\
      & ExtraTrees & 11.55  & \uline{23.60} & 60.91  & 46.20  & 78.33  & \uline{59.07} & 46.61  & 3.2 \\
      & LightGBM & 11.43  & 16.20  & 54.28  & 35.70  & 72.06  & 47.56  & 39.54  & 5.7 \\
      & STAND & \textbf{24.47} & \textbf{26.54} & 52.79  & 42.82  & \textbf{80.66} & \textbf{60.09} & \uline{47.89} & \textbf{2.5} \\
\bottomrule
\end{tabular}%
}
  \label{tab:sp3_comparison}%
\end{table}%

\subsubsection{Event-Level Few-Shot Label Efficiency}\label{sec:fewshot_efficiency}
\begin{figure}[pos=htb]
\centering
\includegraphics[width=0.95\linewidth]{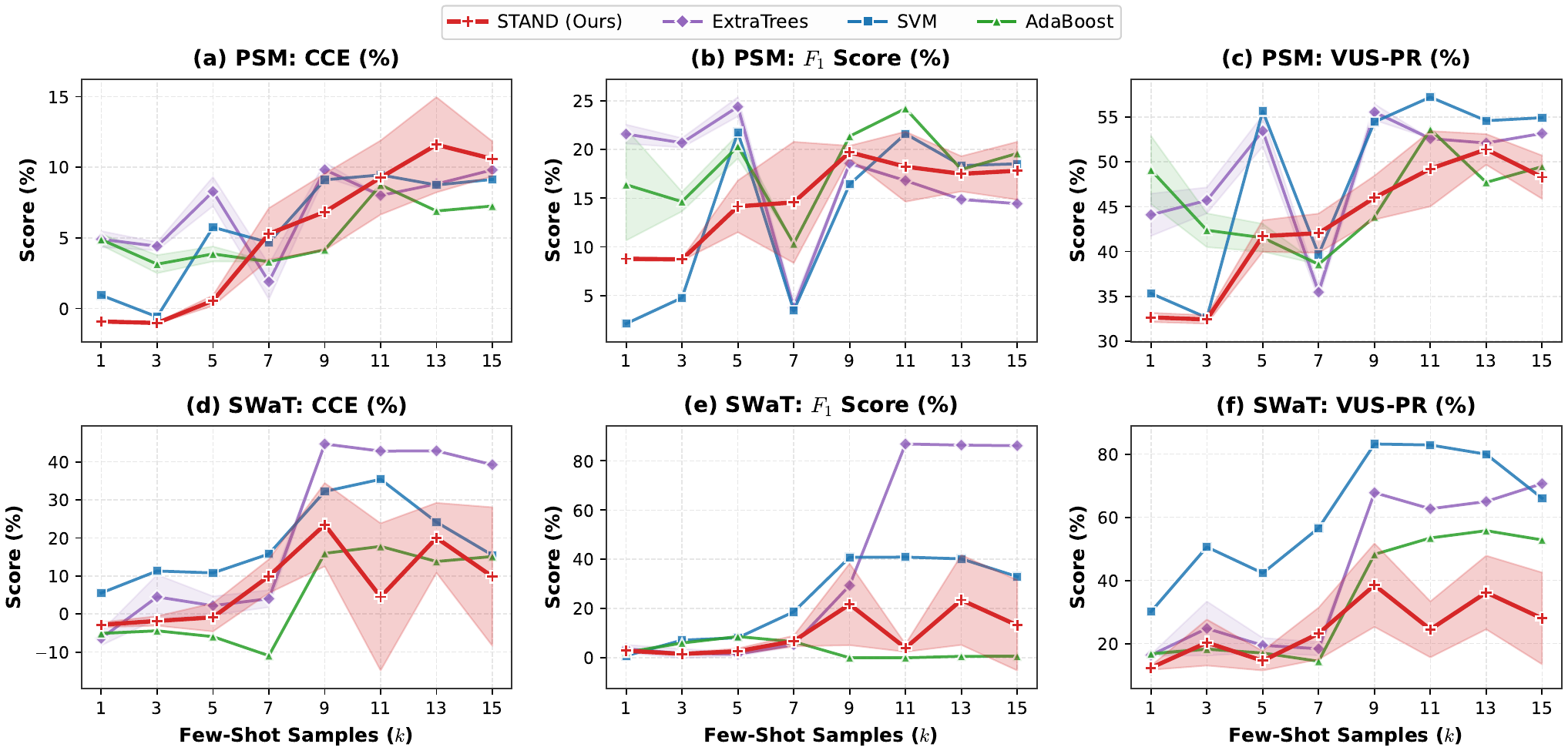}
\caption{\added{Event-level few-shot label efficiency of supervised anomaly detection models across varying numbers of observed failure episodes ($k \in \{1, 3, \dots, 15\}$) on PSM and SWaT benchmarks. Solid lines and shaded bands represent empirical means and standard deviations across experimental runs. A decisive performance transition occurs at $k = 5 \sim 7$ episodes. Detailed tabulated metrics are reported in Supplementary Table~\ref{tab:fewshot_impact}.}}
\label{fig:fewshot_scaling}
\end{figure}

\added{To investigate the minimal volume of historical supervisory annotations required to surpass unsupervised baselines, we conduct an event-level few-shot evaluation across the PSM and SWaT benchmarks. Rather than utilizing arbitrary timestamp fractions, we construct training partitions parameterized by $k \in \{1, 3, 5, 7, 9, 11, 13, 15\}$ discrete, contiguous anomaly episodes alongside their surrounding operational context, replicating practical industrial conditions where only a handful of failure incidents have been archived. Figure~\ref{fig:fewshot_scaling} traces performance trajectories across metrics as the observed anomaly budget $k$ scales, while complete numerical measurements are documented in Supplementary Table~\ref{tab:fewshot_impact}.}

\added{The empirical trajectories reveal a clear break-even threshold around $k = 5 \sim 7$ discrete events. Under extreme low-shot conditions ($k \le 3$), discriminative models yield $\text{CCE} < 0$ due to insufficient support for characterizing multivariate anomaly distributions. However, once $k$ reaches $5$ to $7$ observed failure episodes, \stand's performance scales steeply: CCE rises from $-0.80$ ($k=1$) to $8.03$ ($k=7$) on PSM and from $-1.81$ ($k=1$) to $16.70$ ($k=7$) on SWaT, overtaking competitive unsupervised deep learning baselines on both benchmarks. As $k$ further expands toward $13 \sim 15$ events, \stand converges toward full-supervision performance (scoring CCE of $15.36$ on PSM and $30.64$ on SWaT), while tree ensembles like ExtraTrees and SVM similarly demonstrate substantial label efficiency. These findings confirm that exhaustive historical labeling is not a prerequisite for supervised viability; observing merely $5$ to $7$ archived failure events is sufficient to surpass complex unsupervised alternatives.}

\subsection{Robustness to Operational Label Corruption}\label{sec:noise_analysis}
While Section~\ref{sec:sup_gain} examines model behavior under clean annotations with constrained event budgets, real-world industrial deployments additionally encounter \textbf{in-sample label corruption} due to logging latency, sensor miscalibration, and human annotator disagreement. To evaluate whether supervised models remain viable when training annotations contain explicit label errors, we examine the resilience of STAD models against two synthetic corruption mechanisms on the PSM benchmark:
\begin{itemize}
    \item \textbf{Point Noise (Symmetric Timestamp Inversion)}: Sporadic logging mistakes where individual timestamp labels are randomly inverted ($0 \leftrightarrow 1$) with independent probability $\rho \in [0.0, 0.10]$.
    \item \textbf{Contiguous Segment Noise (False-Positive Operational Blocks)}: Systematic misclassification where contiguous blocks of nominal timestamps are inverted to anomalous until reaching the overall budget $\rho$, simulating false alarm intervals or misidentified operational maintenance cycles.
\end{itemize}
To ensure statistical reliability across distinct corruption patterns, each noise configuration is evaluated across multiple repeated runs with independent random corruption realizations.

\begin{figure}[pos=htb]\centering
\includegraphics[width=0.95\linewidth]{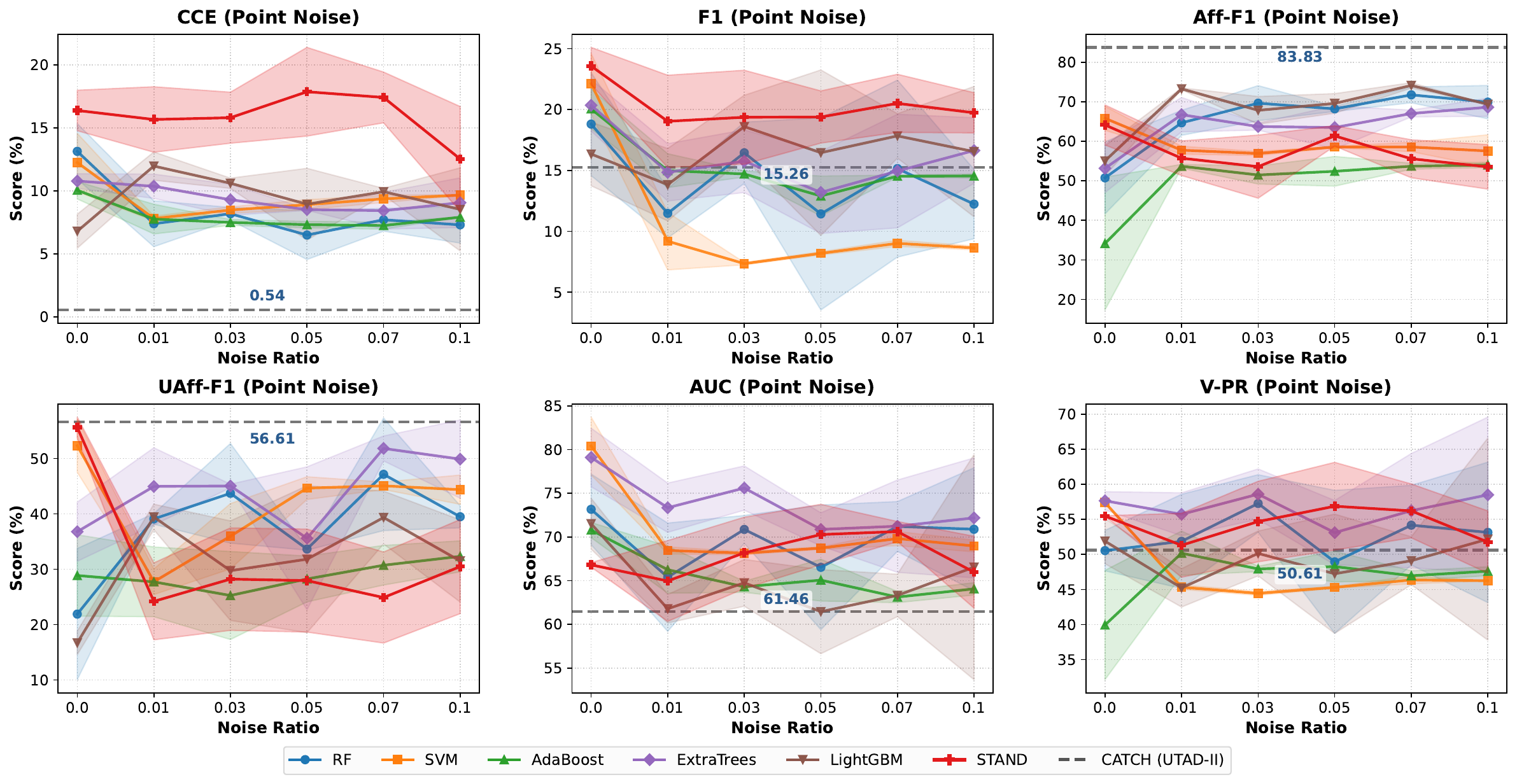}
\caption{Robustness of supervised anomaly detectors under varying ratios of Point Noise on the PSM benchmark. Solid curves depict the empirical mean across multiple repeated runs with distinct random corruption realizations, and shaded error bands denote the standard deviation ($\pm 1\sigma$). The horizontal dashed line indicates the performance of CATCH \citep{catch}, an unsupervised UTAD-II baseline evaluated on the nominal sequence. Complete results for contiguous Segment Noise are reported in Supplementary Figure \ref{fig:supp_segment_noise}.}\label{fig:noise_comparison}
\end{figure}

We systematically examine noise corruption ratios up to $\rho=0.10$ alongside CATCH \citep{catch}, an unsupervised UTAD-II architecture that operates without annotations and serves as an invariant horizontal reference. Figure \ref{fig:noise_comparison} plots model performance under Point Noise across six evaluation metrics, where solid curves and shaded bands denote empirical means and standard deviations ($\pm 1\sigma$) across repeated runs (with detailed results for contiguous Segment Noise provided in Supplementary Section~\ref{app:noise} and Supplementary Figure \ref{fig:supp_segment_noise}). Supervised models, particularly \stand, demonstrate pronounced stability against point corruptions. On prediction certainty (CCE), \stand consistently records the highest certainty across all tested corruption ratios ($12.6 \sim 17.9$), substantially exceeding the unsupervised reference score of $0.54$. Similarly, on F1, \stand maintains superior performance throughout the entire corruption range up to $\rho=0.10$, registering scores between $19.0$ and $23.5$ in comparison to the CATCH reference level of $15.26$. On global discriminative ranking metrics (AUC-ROC and VUS-PR), \stand and tree ensembles firmly exceed the corresponding unsupervised reference thresholds across all tested noise ratios.

Under contiguous Segment Noise (detailed in Supplementary Figure \ref{fig:supp_segment_noise}), continuous mislabeled blocks alter local transition boundaries more severely, causing localized affiliation metrics such as UAff-F1 to cross below CATCH when the corruption ratio reaches $\rho \approx 0.03 \sim 0.05$. Nevertheless, \stand preserves superior prediction certainty (CCE) and discriminative ranking throughout the entire noise spectrum up to $\rho=0.10$. This empirical resilience indicates that sequential recurrent modeling in \stand provides an inductive temporal smoothing effect: because recurrent state transitions integrate context over temporal windows, the model resists overfitting to isolated label corruptions, thereby preserving dependable anomaly discrimination in noise-prone deployments. We further discuss the implications of asymmetric false-negative label noise (missing positive annotations in training) in Section~\ref{sec:limitations}.

\subsection{Effectiveness Analysis of \stand}\label{sec:effectiveness_analysis}
To validate the architectural design of \stand and assess the individual contribution of each component, we conduct an extensive ablation study across all five benchmark datasets. The evaluation examines two primary dimensions: first, the necessity of temporal sequence modeling and the impact of bidirectional context versus causal streaming; second, the comparative effectiveness of different sequential encoder architectures. The consolidated empirical results are reported in Table \ref{tab:ablation_all}.

\subsubsection{Component Necessity and Streaming Causality Analysis}\label{sec:component_ablation}
We examine three configurations to decouple the impact of sequence modeling from directional context: STAND-Pointwise, which replaces the recurrent temporal encoder with an identity mapping and thus reduces the model to a pointwise multilayer perceptron; STAND-Causal, which retains the LSTM encoder but restricts hidden states exclusively to the forward direction ($\overrightarrow{h}_t$), enforcing strict temporal causality for real-time streaming deployments; and the default \stand, which incorporates bidirectional recurrent encoding ($\overleftrightarrow{h}_t$) for offline diagnostic auditing.

As presented in Table \ref{tab:ablation_all}, the experimental findings confirm the indispensable role of sequential representation in multivariate anomaly detection. When temporal modeling is removed in STAND-Pointwise, detection capability degrades across all five benchmarks. Most prominently, the average CCE score drops from $38.50$ to $10.48$, corresponding to a $72.8\%$ decline in prediction consistency, while the average F1 score decreases from $24.32$ to $16.35$ and AUC-ROC declines from $80.89$ to $65.36$. This substantial drop demonstrates that isolated timestamp-level embeddings without historical dynamics are insufficient to separate genuine anomalies from routine operating state changes.

Introducing unidirectional recurrence in STAND-Causal largely restores detection performance, achieving an average CCE of $28.67$, an F1 score of $23.32$, and an AUC-ROC of $80.61$. This result carries significant practical implications: in operational streaming environments where future timestamps cannot be observed, STAND-Causal serves as an effective, strictly causal detector with strong consistency and event localization capabilities. When bidirectional hidden states are enabled in the default \stand model for offline auditing, the average CCE increases further to $38.50$ and unbiased event localization (UAff-F1) improves to $58.22$, which is $9.44$ percentage points higher than STAND-Causal. This supplementary gain indicates that complex anomalies frequently exhibit characteristic post-incident recovery patterns, which the backward pass effectively leverages to refine event boundaries.

\begin{table}[htbp]
  \centering
  \caption{Ablation results of \stand across five datasets. STAND-Pointwise removes the temporal encoder, reducing to a pointwise MLP; STAND-Causal employs a unidirectional LSTM for strict online streaming; STAND-Attn and STAND-GRU substitute the BiLSTM with Multi-Head Attention and Bidirectional GRU, respectively; \stand (Default) is the full bidirectional model. \textbf{Bold} and \uline{underlining} denote the best and second-best performance within each dataset block.}
  \setlength{\tabcolsep}{3.5pt}
  \renewcommand{\arraystretch}{1.1}
  \resizebox{0.82\linewidth}{!}{
    \begin{tabular}{c|l|cc|cccccc}
    \toprule
    \textbf{Dataset} & \textbf{Variant} & \textbf{Encoder} & \textbf{Bidir.} & \textbf{CCE} & \textbf{F1} & \textbf{Aff-F1} & \textbf{UAff-F1} & \textbf{AUC-ROC} & \textbf{VUS-PR} \\
    \midrule
    \multirow{5}{*}{PSM} & STAND-Pointwise & None (MLP) & \ding{56}  &  4.51 & 13.59 & \textbf{78.69} & \uline{48.71} & 59.21 & 44.24 \\
     & STAND-Causal  & Uni-LSTM   & \ding{56}  & \textbf{16.39} & \textbf{23.58} & \uline{64.18} & \textbf{55.61} & \textbf{66.81} & \textbf{55.47} \\
     & STAND-Attn    & Attention  & \ding{52}  &  7.02 & 10.28 & 59.97 & 15.55 & 55.88 & 39.56 \\
     & STAND-GRU     & Bi-GRU     & \ding{52}  & \uline{11.58} & \uline{15.63} & 61.55 & 35.79 & \uline{62.70} & \uline{46.98} \\
     & \stand (Default)       & Bi-LSTM    & \ding{52}  & \textbf{16.39} & \textbf{23.58} & \uline{64.18} & \textbf{55.61} & \textbf{66.81} & \textbf{55.47} \\
    \midrule
    \multirow{5}{*}{SWaT} & STAND-Pointwise & None (MLP) & \ding{56}  & 20.37 & 41.01 & \textbf{82.26} & \textbf{77.34} & 74.46 & 57.22 \\
     & STAND-Causal  & Uni-LSTM   & \ding{56}  & 33.71 & 39.65 & 75.03 & 61.06 & 86.93 & 67.92 \\
     & STAND-Attn    & Attention  & \ding{52}  & \uline{47.36} & \uline{41.30} & 43.29 & 43.18 & 90.02 & \textbf{85.59} \\
     & STAND-GRU     & Bi-GRU     & \ding{52}  & 25.71 & \textbf{41.50} & 34.59 & 34.43 & \uline{90.46} & \uline{80.67} \\
     & \stand (Default)       & Bi-LSTM    & \ding{52}  & \textbf{51.97} & 36.88 & \uline{76.94} & \uline{69.73} & \textbf{90.68} & 79.62 \\
    \midrule
    \multirow{5}{*}{WADI} & STAND-Pointwise & None (MLP) & \ding{56}  &  3.04 & 13.63 & \uline{80.02} & 56.29 & 67.97 & 13.64 \\
     & STAND-Causal  & Uni-LSTM   & \ding{56}  & 15.47 & \uline{29.65} & 77.86 & 36.53 & \uline{75.48} & \uline{27.23} \\
     & STAND-Attn    & Attention  & \ding{52}  & 12.39 & 21.05 & \textbf{85.34} & \textbf{63.32} & 71.92 & 18.93 \\
     & STAND-GRU     & Bi-GRU     & \ding{52}  & \uline{16.93} & 18.66 & 79.44 & \uline{62.86} & \textbf{81.06} & 19.38 \\
     & \stand (Default)       & Bi-LSTM    & \ding{52}  & \textbf{26.70} & \textbf{37.33} & 78.10 & 55.51 & 73.17 & \textbf{39.71} \\
    \midrule
    \multirow{5}{*}{Swan} & STAND-Pointwise & None (MLP) & \ding{56}  &  1.72 & 10.31 & \textbf{31.20} & 17.45 & 48.98 & 91.51 \\
     & STAND-Causal  & Uni-LSTM   & \ding{56}  & \uline{33.26} & \textbf{17.17} & 27.39 & \uline{26.95} & 77.27 & \textbf{96.69} \\
     & STAND-Attn    & Attention  & \ding{52}  & \textbf{33.32} & 15.95 & \uline{30.19} & \textbf{28.77} & \textbf{81.60} & 95.37 \\
     & STAND-GRU     & Bi-GRU     & \ding{52}  & 31.16 & \uline{16.06} & 25.81 & 24.70 & \uline{80.94} & \uline{95.46} \\
     & \stand (Default)       & Bi-LSTM    & \ding{52}  & \uline{33.26} & \textbf{17.17} & 27.39 & \uline{26.95} & 77.27 & \textbf{96.69} \\
    \midrule
    \multirow{5}{*}{Water} & STAND-Pointwise & None (MLP) & \ding{56}  & 22.79 &  3.21 & \uline{86.62} & \uline{68.77} & 76.20 &  4.10 \\
     & STAND-Causal  & Uni-LSTM   & \ding{56}  & \uline{44.53} & \uline{ 6.52} & 84.58 & 63.74 & \textbf{96.55} & \textbf{31.13} \\
     & STAND-Attn    & Attention  & \ding{52}  & 27.03 &  1.44 & 86.54 & 68.76 & 79.50 &  1.17 \\
     & STAND-GRU     & Bi-GRU     & \ding{52}  & 14.03 &  2.87 & 85.79 & 66.71 & 81.80 &  3.21 \\
     & \stand (Default)       & Bi-LSTM    & \ding{52}  & \textbf{64.19} & \textbf{ 6.63} & \textbf{92.21} & \textbf{83.33} & \uline{96.53} & \uline{20.37} \\
    \midrule
    \midrule
    \multirow{5}{*}{\textbf{Average}} & STAND-Pointwise & None (MLP) & \ding{56}  & 10.48 & 16.35 & \textbf{71.76} & \uline{53.71} & 65.36 & 42.14 \\
     & STAND-Causal  & Uni-LSTM   & \ding{56}  & \uline{28.67} & \uline{23.32} & 65.81 & 48.78 & \uline{80.61} & \uline{55.69} \\
     & STAND-Attn    & Attention  & \ding{52}  & 25.42 & 18.00 & 61.07 & 43.92 & 75.78 & 48.12 \\
     & STAND-GRU     & Bi-GRU     & \ding{52}  & 19.88 & 18.95 & 57.44 & 44.90 & 79.39 & 49.14 \\
     & \stand (Default)       & Bi-LSTM    & \ding{52}  & \textbf{38.50} & \textbf{24.32} & \uline{67.76} & \textbf{58.22} & \textbf{80.89} & \textbf{58.37} \\
    \bottomrule
    \end{tabular}
  }
  \label{tab:ablation_all}
\end{table}

\subsubsection{Impact of Temporal Encoder Architecture}\label{sec:encoder_ablation}
To assess whether alternative sequence models offer advantages over the recurrent baseline, we compare the default BiLSTM against a Multi-Head Attention module (STAND-Attn) and a Bidirectional GRU (STAND-GRU). As detailed in Table \ref{tab:ablation_all}, the default BiLSTM architecture attains superior overall performance across the evaluated benchmarks, scoring an average CCE of $38.50$ compared to $25.42$ for Attention and $19.88$ for GRU, alongside an average F1 score of $24.32$ versus $18.00$ and $18.95$, respectively.

These comparative results illuminate the architectural trade-offs in supervised sequence modeling. Self-attention mechanisms introduce quadratic parameter scaling and lack strong temporal inductive bias, which increases their risk of overfitting to localized anomaly tokens and weakens generalization on unseen test streams. In contrast, while the GRU eliminates the dedicated memory cell to reduce parameter count, this simplification limits its capacity to maintain long-range context across high-dimensional industrial telemetry, leading to lower localized detection scores such as an average VUS-PR of $49.14$ compared to $58.37$ for \stand. By contrast, the bidirectional LSTM strikes an effective balance between sequential expressive capacity and parameter efficiency, yielding stable prediction certainty and precise anomaly boundary delineation.

\subsubsection{Cross-Paradigm Comparison with Matched Backbones}\label{sec:matched_backbone}
\added{To rigorously disentangle the empirical impact of the supervisory learning objective from sequential encoder capacity, we compare identical temporal architectures trained under two distinct paradigms: unsupervised reconstruction via mean squared error without anomaly feedback (STAND-Recon, Panel~A), and discriminative classification via binary cross-entropy with historical labels (\stand, Panel~B). As summarized in Table~\ref{tab:matched_backbone_summary} (with per-benchmark breakdowns across all five datasets detailed in Supplementary Section~\ref{app:matched_backbone}), the supervisory learning objective exerts a decisive influence on anomaly detection capability.}

\begin{table}[pos=t]
\centering
\caption{\added{Cross-paradigm comparison with matched sequential backbones averaged across all five benchmark datasets. Panel~A reports unsupervised reconstruction ($0$ anomaly labels, MSE loss); Panel~B reports supervised classification (full training labels, BCE loss); Panel~C reports the net supervisory gain ($\Delta = \text{Panel~B} - \text{Panel~A}$). Detailed per-dataset evaluations are provided in Supplementary Section~\ref{app:matched_backbone}.}}
\label{tab:matched_backbone_summary}
\resizebox{0.92\columnwidth}{!}{%
\begin{tabular}{lcccccc}
\toprule
\textbf{Encoder Architecture} & \textbf{CCE} & \textbf{F1} & \textbf{Aff-F1} & \textbf{UAff-F1} & \textbf{AUC-ROC} & \textbf{VUS-PR} \\
\midrule
\multicolumn{7}{l}{\textit{Panel A: Unsupervised Reconstruction (0 labels, MSE)}} \\
\midrule
Bi-LSTM (STAND-Recon)   &  1.13 & 13.55 & \textbf{64.18} & 21.97 & 61.19 & 40.64 \\
Attention (Attn-Recon)  &  1.08 & 13.59 & 65.04 & 21.07 & 60.63 & 40.31 \\
Bi-GRU (GRU-Recon)      &  1.11 & 13.70 & 63.88 & 24.54 & 59.16 & 40.10 \\
\midrule
\multicolumn{7}{l}{\textit{Panel B: Supervised Classification (Full labels, BCE)}} \\
\midrule
Bi-LSTM (\stand)        & \textbf{38.50} & \textbf{24.32} & 67.76 & \textbf{58.22} & \textbf{80.89} & \textbf{58.37} \\
Attention (STAND-Attn)  & \textbf{25.42} & \textbf{18.00} & 61.07 & \textbf{43.92} & \textbf{75.78} & \textbf{48.12} \\
Bi-GRU (STAND-GRU)      & \textbf{19.88} & \textbf{18.95} & 57.44 & \textbf{44.90} & \textbf{79.39} & \textbf{49.14} \\
\midrule
\multicolumn{7}{l}{\textit{Panel C: Net Supervisory Gain ($\Delta = \text{Panel B} - \text{Panel A}$)}} \\
\midrule
$\Delta$ (Bi-LSTM)      & +37.37 & +10.77 & +3.58 & +36.25 & +19.70 & +17.73 \\
$\Delta$ (Attention)    & +24.34 & +4.41  & -3.97 & +22.85 & +15.15 & +7.81  \\
$\Delta$ (Bi-GRU)       & +18.77 & +5.25  & -6.44 & +20.36 & +20.23 & +9.04  \\
\bottomrule
\end{tabular}%
}
\end{table}

\added{Across all three sequence encoders, unsupervised reconstruction constrains CCE to a narrow range ($1.08 \sim 1.13$), reflecting the intrinsic limitation of reconstruction residuals where normal operational variability often obscures failure boundaries. In contrast, under identical sequential inductive biases, optimizing classification objectives delivers substantial performance leaps: CCE surges by $+18.77$ to $+37.37$, UAff-F1 improves by $+20.36$ to $+36.25$ percentage points, and AUC-ROC increases by $+15.15$ to $+20.23$. These controlled comparisons substantiate that the primary performance bottleneck in unsupervised time series anomaly detection originates from the surrogate learning objective rather than the expressiveness of the underlying sequence encoder.}

\begin{figure}[pos=htb]
    \centering
    \includegraphics[width=\linewidth]{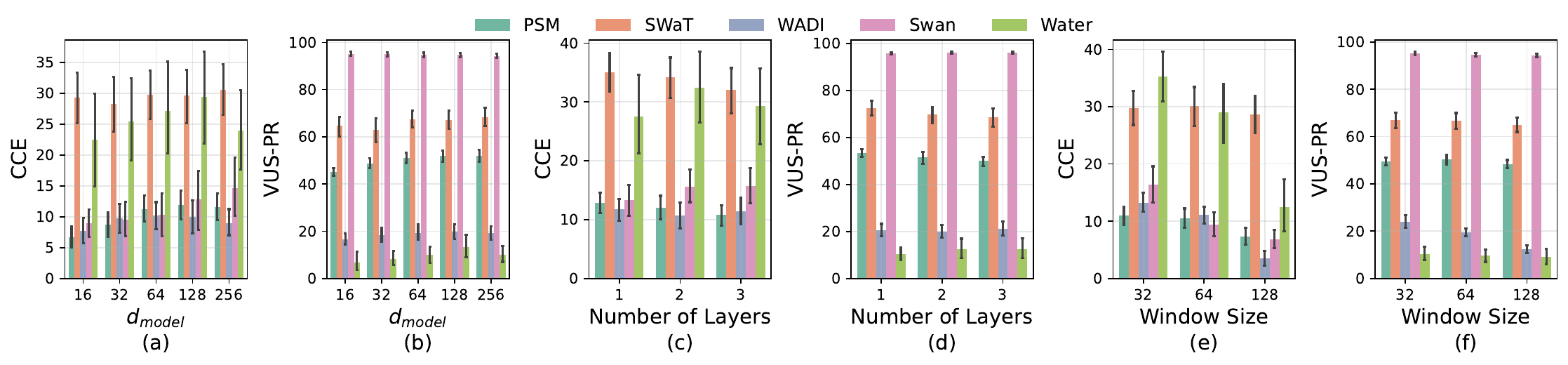}
    \vspace{-8pt}  
    \caption{Sensitivity analysis of model performance with respect to three key hyperparameters on five datasets. Plots (a) and (b) present the impact of model dimension $d_{\text{model}}$ on CCE and VUS-PR, respectively; (c) and (d) illustrate the effect of the number of TEM layers; (e) and (f) depict the influence of temporal window size. Error bars denote 95\% confidence intervals across hyperparameter configurations and independent runs.}
    \label{fig:sen1}
\end{figure}
\begin{figure}[pos=htb]
    \centering
    \includegraphics[width=\linewidth]{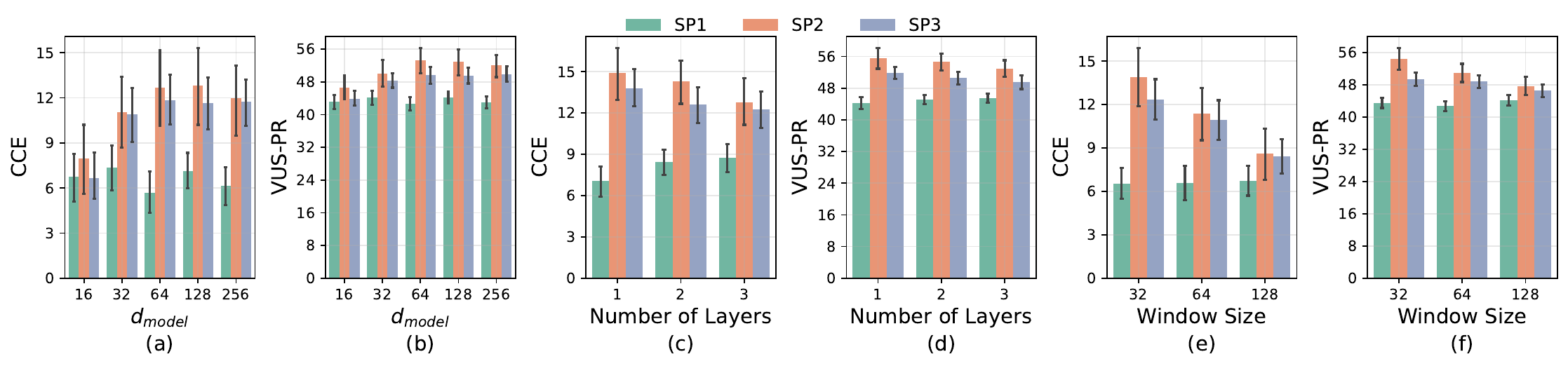}
    \vspace{-8pt}  
    \caption{Sensitivity analysis of model performance with respect to three key hyperparameters across three subsets of the PSM dataset, SP1--SP3. Plots (a) and (b) present the impact of model dimension $d_{\text{model}}$ on CCE and VUS-PR, respectively; (c) and (d) illustrate the effect of the number of TEM layers; (e) and (f) depict the influence of temporal window size. Error bars denote 95\% confidence intervals across hyperparameter configurations and independent runs.}
    \label{fig:sen2}
\end{figure}

\subsection{Sensitivity Analysis of STAND}\label{sec:sensitivity}
To investigate the influence of hyperparameters on \stand under out-of-sample evaluation, we analyze model sensitivity on CCE and VUS-PR across diverse parameter settings on all benchmark datasets and the PSM subsets. Rather than freezing remaining hyperparameters to arbitrary baseline values, each hyperparameter level is evaluated via marginal sensitivity analysis across the configuration grid (varying temporal window sizes $W \in \{32, 64, 128\}$ and number of TEM layers $L \in \{1, 2, 3\}$) across independent runs with distinct random initializations. As illustrated in Figures \ref{fig:sen1} and \ref{fig:sen2}, where bars and error bars indicate the marginal sample mean and 95\% confidence intervals (with complete multi-metric evaluations provided in Supplementary Section~\ref{app:sen}), the sensitivity patterns of \stand exhibit high consistency across configurations.

Regarding model dimension $d_{\text{model}}$, scaling from $16$ to $128$ steadily enhances both prediction certainty on CCE and event-level discriminative power on VUS-PR, with performance stabilizing between $128$ and $256$. This upward trend demonstrates that projecting multivariate telemetry into a sufficiently expressive latent space improves the separation of subtle anomalies from complex normal operating states.

In terms of recurrent depth, shallow architectures comprising one or two layers consistently deliver the most dependable detection accuracy. When depth increases to three layers, performance experiences a mild decline across the majority of datasets, indicating that deeper recurrent stacks increase optimization variance without contributing necessary representational benefits in this setting.

With respect to temporal receptive fields, a compact window size of $32$ distinctly outperforms larger spans of $64$ and $128$. Expanding the temporal horizon introduces extraneous normal observations and dilutes the localized boundaries of transient anomalies. Consequently, allocating parameter capacity to latent channel representation proves far more effective than deepening recurrent layers or inflating the temporal input window.

\subsection{Visualization Analysis}\label{sec:vis}
\begin{figure}[pos=htb]
    \centering
    \includegraphics[width=0.85\linewidth]{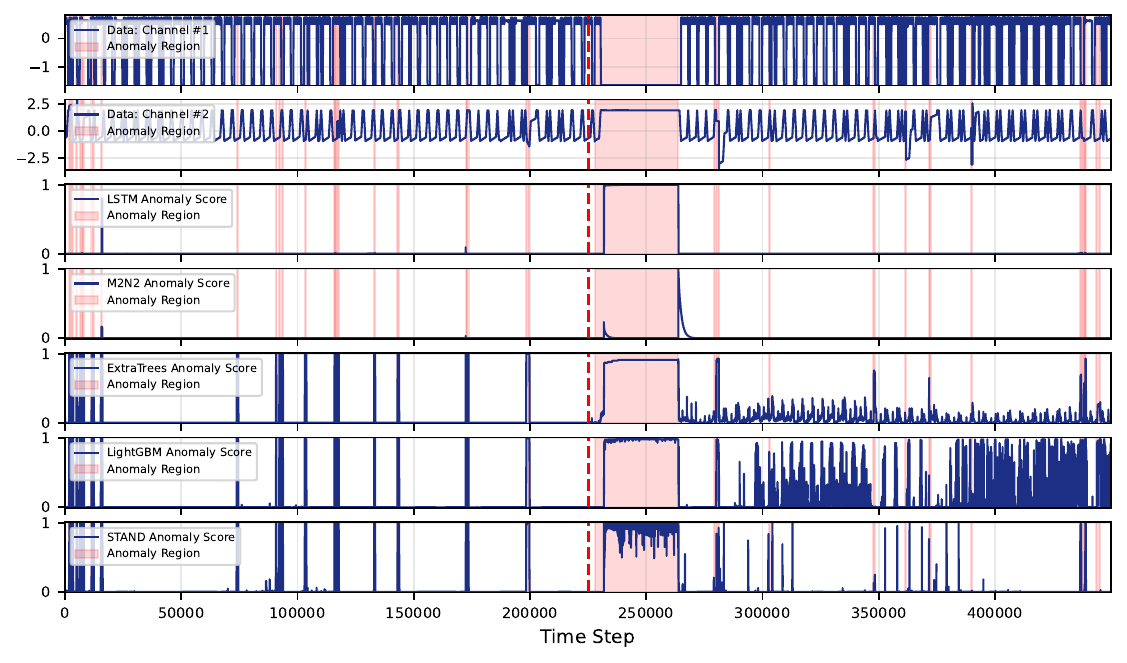}
    \caption{Anomaly score trajectories of representative models across the complete SWaT sequence. The red vertical dashed line separates the training prefix from the unseen test sequence, with ground-truth anomaly regions shaded in light red.}
    \label{fig:vis}
\end{figure}
\begin{figure}[pos=htb]
    \centering
    \includegraphics[width=0.85\linewidth]{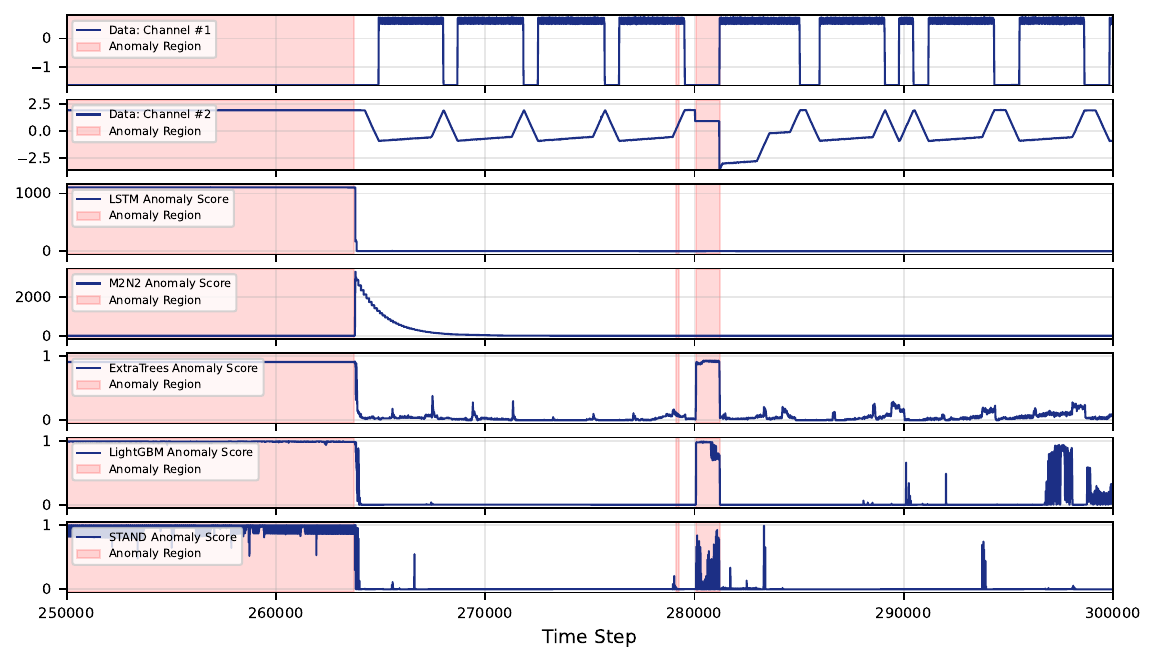}
    \caption{Anomaly score trajectories on an unseen test segment from timestamp 250k to 300k, showing boundary behavior on sustained and transient anomalies.}
    \label{fig:vis1}
\end{figure}
\begin{figure}[pos=htb]
    \centering
    \includegraphics[width=0.85\linewidth]{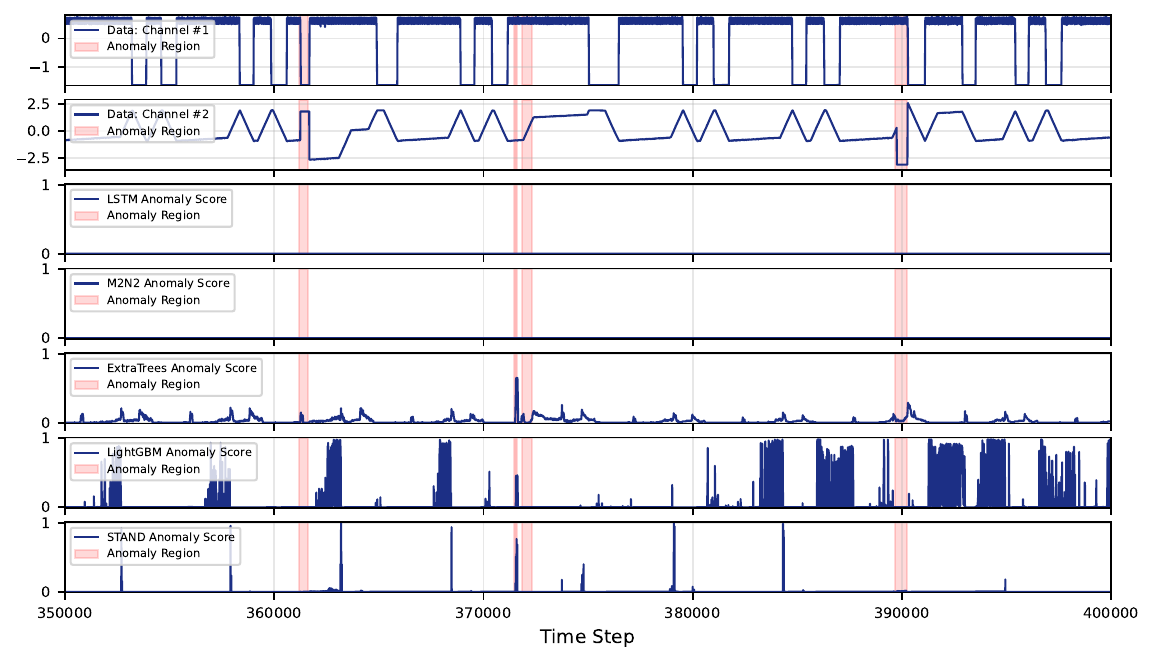}
    \caption{Anomaly score trajectories on an unseen test segment from timestamp 350k to 400k, showing detection performance, transition false alarms, and a missed anomaly under distinct abnormal dynamics.}
    \label{fig:vis2}
\end{figure}

To examine practical detection behavior, we evaluate the anomaly score trajectories of representative unsupervised models, LSTM and M2N2, alongside supervised models: ExtraTrees, LightGBM, and \stand on the SWaT benchmark. In Figures \ref{fig:vis}--\ref{fig:vis2}, the top two panels display two representative telemetry channels: Channel \#1 (FIT101, inflow flow rate) and Channel \#2 (LIT101, tank water level), illustrating contrasting switching and cyclical dynamics. The remaining panels present the anomaly scores produced by each model, with ground-truth anomalies shaded in light red.

Figure \ref{fig:vis} presents a macroscopic view over the entire sequence of 450k timestamps, where the red dashed line marks the boundary between the training prefix and the unseen test sequence. Over the complete sequence, the unsupervised models LSTM and M2N2 remain near zero across most intervals, registering noticeable scores only during the prolonged anomaly in the first half of the test sequence while missing most other anomalies. In contrast, the supervised models generate distinct responses during anomalous intervals across both training and test data, while exhibiting different background false alarm behaviors. A qualitative inspection focusing on the training prefix is provided in Supplementary Section~\ref{app:vis}.


To examine detection dynamics on unseen data in closer detail, Figure \ref{fig:vis1} presents the segment from timestamp 250k to 300k. During the extended anomaly in the early part of this window, both unsupervised and supervised models produce elevated scores. However, their behaviors when the anomaly ends differ noticeably. Following the anomaly offset, M2N2 displays a prolonged decay that requires thousands of time steps to return to zero. In contrast, \stand drops immediately back to baseline, aligning closely with the true end of the event. During the subsequent shorter anomaly near timestamp 280k, the supervised models register clear detections, whereas LSTM and M2N2 show negligible response. In normal intervals, LightGBM exhibits dense false alarms toward the end of this window, whereas \stand maintains a quiet baseline with few isolated spikes.

Figure \ref{fig:vis2} illustrates a subsequent unseen test segment from timestamp 350k to 400k, depicting both successful detections and model limitations. Near the middle of this segment, \stand and ExtraTrees detect the anomalous interval, whereas the unsupervised baselines remain silent. However, during a subsequent anomaly where Channel \#2 undergoes a sharp downward drop, \stand outputs a score near zero and fails to trigger an alarm. This failure shows that when an anomaly exhibits behavioral dynamics absent from the training labels, discriminative models can fail to generalize to the new pattern. Unsupervised models also miss this event, indicating the general difficulty of capturing this shift. Additionally, \stand exhibits occasional single-step spikes during sudden switching transitions in Channel \#1. Nevertheless, unlike LightGBM, which triggers extensive blocks of false alarms throughout normal periods, \stand keeps false alarms isolated and infrequent, benefiting from sequential temporal modeling.

\subsection{Computational Efficiency and Resource Profiling}\label{sec:efficiency}
To empirically validate the scalability of \stand and evaluate practical deployment viability, Table \ref{tab:efficiency} profiles parameter counts, memory footprints, and runtime efficiency on the PSM benchmark ($C=25$, batch size $128$). All deep learning architectures are evaluated under a standardized window of $W=100$ for direct comparability, while classical baselines employ their benchmark-recommended configurations ($W=30$ for PCA and $W=3$ for ExtraTrees).

\begin{table}[htbp]
  \centering
  \caption{Empirical computational efficiency and resource consumption on the PSM benchmark ($C=25$, batch size $128$). All deep architectures are evaluated under $W=100$. Peak VRAM represents the maximum GPU memory allocated during training. Inference latency is measured per timestamp. For non-iterative CPU models (PCA and ExtraTrees), training time denotes the total fitting duration on the training split.}
  \label{tab:efficiency}
  \resizebox{\linewidth}{!}{
    \begin{tabular}{lccccccc}
    \toprule
    \textbf{Model} & \textbf{Paradigm} & \textbf{Win. ($W$)} & \textbf{Params (M)} & \textbf{Peak VRAM (MB)} & \textbf{Train Time (s/epoch)} & \textbf{Infer. Latency ($\mu\text{s}$)} & \textbf{Infer. Throughput ($10^3$ timestamps/s)} \\
    \midrule
    PCA & UTAD (CPU) & 30 & --- & ---$^*$ & 30.62$^\dagger$ & 548.0 & 1.8 \\
    AE & UTAD (GPU) & 100 & 0.02 & 17.7 & 1.00 & 10.3 & 96.7 \\
    Anomaly Transformer & UTAD (GPU) & 100 & 4.80 & 5497.1 & 34.21 & 240.0 & 4.2 \\
    CATCH & UTAD (GPU) & 100 & 89.86 & 2231.7 & 33.33 & 7.6 & 132.0 \\
    ExtraTrees & STAD (CPU) & 3 & --- & ---$^*$ & 0.27$^\dagger$ & 1.3 & 798.1 \\
    {STAND (Ours)} & STAD (GPU) & 100 & 0.04 & 86.5 & 1.96 & 8.8 & 113.7 \\
    \bottomrule
    \multicolumn{8}{l}{\footnotesize $^*$Operates entirely on host CPU memory with negligible footprint ($<200$ MB RAM). $^\dagger$Denotes total fitting time on the training split.}
    \end{tabular}%
  }
\end{table}

The empirical measurements substantiate the minimalist design and operational efficiency of \stand. With $0.04\,\text{M}$ parameters and $86.5\,\text{MB}$ peak VRAM, \stand reduces memory overhead by over $96\%$ compared to CATCH ($2231.7\,\text{MB}$) and $98\%$ compared to Anomaly Transformer ($5497.1\,\text{MB}$), easily accommodating memory-constrained edge gateways. In terms of runtime, \stand attains an inference latency of $8.8\,\mu\text{s}$ per timestamp, executing over $27\times$ faster than Anomaly Transformer ($240.0\,\mu\text{s}$), where full-matrix attention projections incur substantial latency penalties. This translates to a high operational throughput of $113.7 \times 10^3$ timestamps/s, confirming that recurrent encoding on practical sliding windows achieves high computational throughput while accommodating high-frequency telemetry processing, whether in causal real-time streaming deployments (STAND-Causal) or rapid offline diagnostic auditing (\stand).

\section{Discussion and Limitations}\label{sec:discussion}

\subsection{Mechanisms and Lifecycle Economics}
The performance advantage of \stand stems from two factors: objective alignment and model parsimony. Unsupervised detectors optimize surrogate losses like reconstruction error or contrastive distance. In industrial telemetry, normal operational shifts often inflate reconstruction residuals without true faults, blurring anomaly boundaries and suppressing prediction certainty (Table~\ref{tab:comparison}, CCE $<6.0$). In contrast, binary cross-entropy trains the model to separate anomalous deviations directly, yielding sharp bimodal probabilities and prompt post-incident recovery (Figure~\ref{fig:vis1}). Furthermore, scarce anomaly feedback favors compact inductive biases over architectural scale. As shown in Table~\ref{tab:ablation_all}, replacing recurrence with multi-head self-attention degrades both CCE and F1, indicating that heavy attention mechanisms overfit on localized anomaly patterns. A single-layer recurrent encoder ($0.04$\,M parameters, Table~\ref{tab:efficiency}) provides effective temporal regularization while remaining computationally lightweight.

These dynamics delineate complementary roles for unsupervised and supervised tools across the monitoring lifecycle. Unsupervised methods remain essential for cold-start exploration on newly commissioned assets without fault history. Over extended operation, however, verified failure records naturally accumulate through routine maintenance logs and automated alarm trips. Harvesting these existing records incurs negligible annotation expense while providing substantial empirical returns. As shown in Table~\ref{tab:sp3_comparison}, observing just $12$ historical events on PSM (subset SP1) allows \stand to outperform deep unsupervised baselines, with detection performance scaling steadily as maintenance archives grow to $35$ events (SP3).

\subsection{Practical Assumptions and Limitations}\label{sec:limitations}
While the empirical results substantiate the utility of supervisory guidance across benchmarks, three practical operational constraints bound the deployment of the supervised paradigm:

\subsubsection{Operational Label Acquisition and Industrial Economics}\label{sec:label_economics}
Our scaling evaluations demonstrate that supervised models become competitive once a small collection of historical failure episodes is observed. In industrial cyber-physical systems, public utility facilities (e.g., SWaT, WADI), and cloud server clusters (e.g., PSM), verified anomaly annotations are not generated by human operators manually scanning millions of raw, high-frequency continuous telemetry timestamps. Instead, verified labels are realistically harvested through three established operational workflows:
\begin{enumerate}
    \item \textit{SCADA alarm trip and emergency safety interlock logs}: Industrial control systems automatically log emergency physical shut-offs, circuit-breaker trips, or valve interlocks when telemetry parameters breach physical safety thresholds, producing unambiguous, timestamped records of severe breakdown windows.
    \item \textit{Post-incident root cause analysis (RCA) records}: Following an operational shutdown, service interruption, or equipment malfunction, engineering teams conduct forensic investigations and record exact failure onset and recovery windows in computerized maintenance management systems (CMMS).
    \item \textit{Routine scheduled maintenance and component replacement orders}: Periodic component servicing and sensor recalibration produce verified historical records of degraded operating states.
\end{enumerate}
Because these operational records are already generated as routine compliance, regulatory, and safety byproducts, retroactively harvesting anomaly intervals from existing maintenance archives incurs near-zero additional manual labeling expense. Furthermore, because severe operational failures are rare, these archives naturally contain only a small, sparse collection of verified failure events (e.g., 12 failure episodes in PSM-SP1, 21 in SWaT, and 21 in Water), closely matching the limited-supervision regimes investigated in our scaling evaluations.

\subsubsection{Historical Selection Bias, Missing Positives, and Open-Set Generalization}\label{sec:selection_bias}
A fundamental limitation of supervised discriminative modeling is its reliance on historical precedence: the model operates under the inductive assumption that future anomalies share underlying physical degradation dynamics (such as sensor decorrelation, gradient saturation, or cyclic phase disruption) with recorded historical events. However, operational failure archives inherently carry \textit{selection bias}: historical logs naturally over-represent known, severe, or easily identifiable failure modes, whereas subtle, nascent, or slow-drifting degradations are frequently omitted (systematic false negatives / missing positive labels in training). 

When encountering zero-precedent, novel anomaly modes absent from historical training logs, discriminative models cannot extrapolate effectively. As illustrated qualitatively in Figure~\ref{fig:vis2}, \stand misses the unobserved downward drift in Channel~\#2 because no corresponding downward failure mode occurred in the training window. For detecting zero-precedent novel anomalies, unsupervised methods retain an indispensable operational advantage by flagging general deviations from nominal dynamics without requiring historical precedent. In contrast, supervised models excel at capturing recurrence patterns of known failure modes with high prediction certainty and sharp boundary delineation. Consequently, UTAD and STAD serve complementary lifecycles rather than mutually exclusive roles: unsupervised detectors are essential for cold-start exploration on newly commissioned assets, while supervised models provide high-certainty operational monitoring as historical failure archives accumulate. Exploring hybrid architectures that couple unsupervised novelty indicators with compact supervised sequence models constitutes a promising direction for future research.

\subsubsection{Real-Time Streaming Latency and Temporal Causality}\label{sec:causality_limitation}
A structural constraint of default bidirectional sequential modeling (\stand) is that computing backward recurrent hidden states ($\overleftarrow{h}_t$) incurs a look-ahead buffering delay of $W$ future timestamps, restricting default \stand to offline diagnostic auditing or delayed verification workflows. For strict zero-latency online streaming deployments, the unidirectional variant \texttt{STAND-Causal} eliminates this look-ahead buffering delay entirely by restricting hidden states exclusively to the forward direction ($\overrightarrow{h}_t$). As shown in Table~\ref{tab:ablation_all}, \texttt{STAND-Causal} preserves competitive detection accuracy ($28.67$ CCE, $23.32$ F1, $80.61$ AUC-ROC) with only a modest drop in localization precision compared to bidirectional \stand, establishing a viable causal supervised baseline for real-time streaming while isolating the empirical cost of unidirectional versus bidirectional temporal encoding.

\section{Conclusion}\label{sec:conclusion}
This paper re-evaluates the prevailing focus on unsupervised architectural complexity in time series anomaly detection. By introducing \stand, a deliberately minimalist supervised baseline, we systematically quantify the empirical trade-offs between supervisory signals and model complexity. Our empirical evaluations across five benchmark datasets indicate that under standard chronological out-of-sample evaluation, a compact discriminative baseline achieves competitive or superior detection performance, prediction certainty, and localized event delineation relative to complex unsupervised architectures across the evaluated benchmarks. Furthermore, controlled label-scaling and few-shot evaluations confirm that these empirical advantages persist even when restricted to sparse historical failure budgets. These findings do not diminish the value of unsupervised methods in exploratory or zero-annotation scenarios; rather, they highlight the substantial empirical return of supervision in practical deployments and demonstrate the diminishing returns of scaling model complexity without supervisory feedback. Consequently, we encourage future TSAD research to balance algorithm-centric design with data-centric label utilization, incorporating supervised baselines into evaluation benchmarks to better reflect operational engineering trade-offs. In future work, we plan to extend this benchmark to weakly-supervised and active learning protocols to systematically determine the minimal labeling effort required across diverse anomaly topologies, \added{as well as explore hybrid open-set architectures that integrate unsupervised novelty indicators for detecting unseen anomalies}.

\section*{Acknowledgments}
This work was supported in part by National Natural Science Foundation of China No. 62572199, 92467109, and the Major Key Project of PCL, China under Grant PCL2025A11 and PCL2025A13.

\bibliographystyle{elsarticle-num}

\bibliography{main}

\vfill

\printcredits






\end{document}